%% file: acl2023.tex
\pdfoutput=1

\documentclass[11pt]{article}

\usepackage[]{ACL2023}

\usepackage{times}
\usepackage{latexsym}
\usepackage{hyperref}
\usepackage{xcolor}
\usepackage{soul}
\usepackage{amsmath}
\usepackage{booktabs}
\usepackage{xcolor}
\usepackage{url}
\usepackage[T1]{fontenc}
\usepackage{longtable}
\usepackage{graphicx}
\usepackage{float}
\usepackage{multirow}
\usepackage{color}
\usepackage{wrapfig}
\usepackage{booktabs}
\usepackage{amssymb}
\usepackage{subfigure}
\usepackage{bbding}
\usepackage{listings}
\usepackage{pifont}
\usepackage{wasysym}
\usepackage{CJKutf8}
\usepackage{utfsym}
\usepackage{fontawesome}
\usepackage{makecell}

\newcommand{\okmark}{{\textbf{\textcolor[rgb]{0.1, 0.5, 0.1}{[$\checkmark$]}}}}

\newcommand{\bblue}[1]{{\textbf{\color{blue}{#1}}}}
\usepackage[T1]{fontenc}

\newtheorem{theorem}{Theorem}[section]

\usepackage[utf8]{inputenc}

\usepackage{microtype}

\usepackage{inconsolata}

\title{Large Language Models Need Holistically Thought in Medical Conversational QA}

\author{Yixuan Weng$^1$\thanks{ $\quad$These authors contribute this work equally.}, Bin Li$^{*3}$, Fei Xia$^{1,2}$, Minjun Zhu$^{1,2}$, Bin Sun$^{*3}$, Shizhu He$^{1,2}$, Kang Liu$^{1,2}$, Jun Zhao$^{1,2}$ \\
	$^1$ National Laboratory of Pattern Recognition,, Institute of Automation, CAS \\
	$^2$ School of Artificial Intelligence, University of Chinese Academy of Sciences\\
 	$^3$ College of Electrical and Information Engineering, Hunan University\\
\texttt{wengsyx@gmail.com,libincn@hnu.edu.cn, \{shizhu.he, kliu, jzhao\}@nlpr.ia.ac.cn}
}
\begin{document}
\maketitle

\begin{abstract}
The medical conversational question answering (CQA) system aims at providing a series of professional medical services to improve the efficiency of medical care. Despite the success of large language models (LLMs) in complex reasoning tasks in various fields, such as mathematics, logic, and commonsense QA, they still need to improve with the increased complexity and specialization of the medical field. This is because medical CQA tasks require not only strong medical reasoning, but also the ability to think broadly and deeply. In this paper, to address these challenges in medical CQA tasks that need to be considered and understood in many aspects, we propose the Holistically Thought (\textbf{HoT}) method, which is designed to guide the LLMs to perform the diffused and focused thinking for generating high-quality medical responses.
The proposed HoT method has been evaluated through automated and manual assessments in three different medical CQA datasets containing the English and Chinese languages. The extensive experimental results show that our method can produce more correctness, professional, and considerate answers than several state-of-the-art (SOTA) methods, manifesting its effectiveness. Our code is in \url{https://github.com/WENGSYX/HoT}.

\end{abstract}

\section{Introduction}

\begin{figure}[h]
\vspace{0.05cm}
  \centering
    \includegraphics[width=0.47\textwidth]{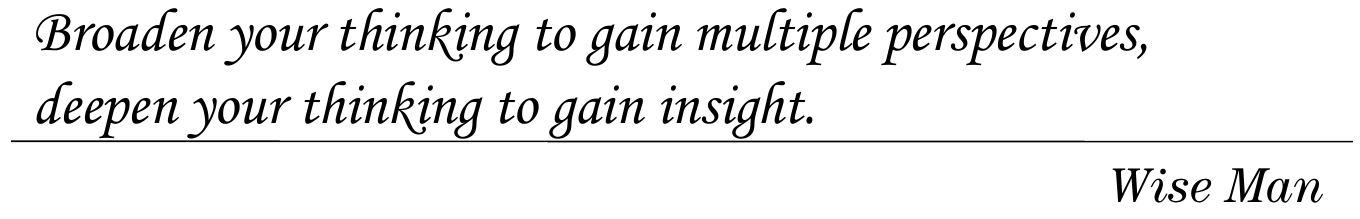}
  \vspace{-0.15cm}
    \label{figure1}
\end{figure}

\noindent Medical research with AI-based techniques is growing rapidly \citep{AurangzebKhan2017ARS,JianfengGao2018NeuralAT,AbdulRehmanJaved2020ACH,li-etal-2022-vpai}. The medical conversational question answering (CQA) task is designed to improve the efficiency of medical care by providing a range of professional medical services for patients \cite{LinXu2019EndtoEndKR}. The development of medical CQA systems has become an important research topic, particularly amid the COVID-19 outbreak \cite{WenmianYang2020OnTG,KunZhou2021CRSLabAO}. Such a system can improve the patient experience during the clinical treatment process by quickly responding to their needs and providing information on symptoms, diagnosis, prescription of medications, and treatment suggestions \cite{AdamPalanica2019PhysiciansPO,chen2021artificial,li2022visualprompt}.

\begin{figure*}[t]
\begin{center}
	\includegraphics[scale=0.45]{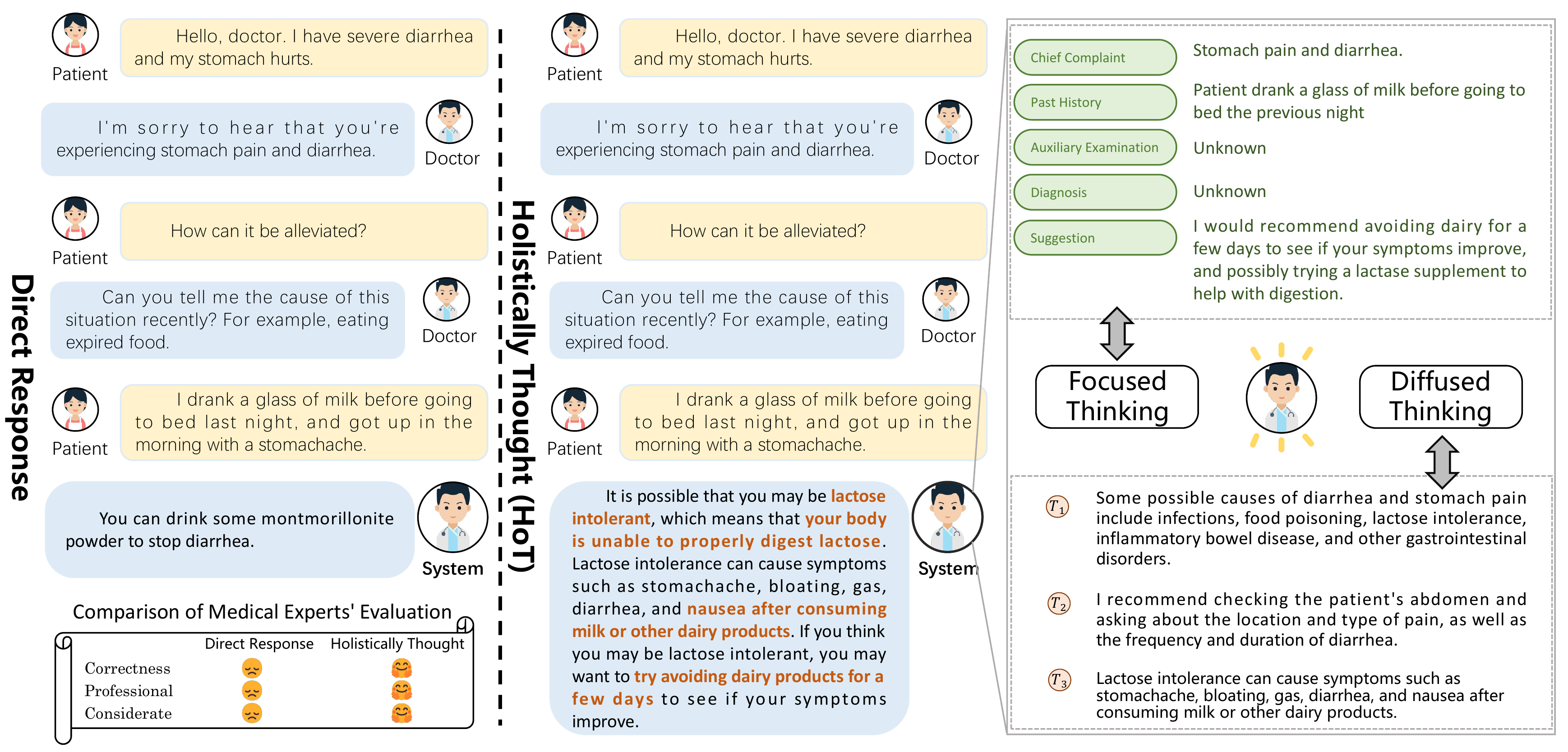}
\end{center}
\vspace{-0.3cm}
\caption{The difference and comparison between direct response generation and consideration of HoT.}
\vspace{-0.3cm}
\label{idea}
\end{figure*}

\noindent The field of medical research incorporating artificial intelligence (AI) techniques has experienced substantial growth in recent years \citep{AurangzebKhan2017ARS,JianfengGao2018NeuralAT,AbdulRehmanJaved2020ACH,li-etal-2022-vpai}. One area of interest is the development of medical conversational question-answering (CQA) systems, which aim to enhance healthcare service efficiency by providing patients with professional support \cite{LinXu2019EndtoEndKR}. These systems hold potential for improving patient experiences during clinical treatment by supplying timely information on symptoms, diagnosis, prescription medications, and treatment recommendations \cite{AdamPalanica2019PhysiciansPO,chen2021artificial,li2022visualprompt}. The significance of medical CQA systems has further intensified amid the COVID-19 pandemic \cite{WenmianYang2020OnTG,KunZhou2021CRSLabAO}.

Although conventional language models such as BERT \cite{JacobDevlin2018BERTPO}, GPT \cite{radford2018improving}, and T5 \cite{raffel2020exploring} have been employed in medical question-answering tasks \cite{QingyangWu2019ImportanceAwareLF,KeLiang2021MKAAS}, their efficacy in the medical domain is limited due to training on insufficient medical datasets \cite{li2021more}. Recent advances in large language models (LLMs) have demonstrated improved performance through in-context learning \cite{dong2022survey} and few-shot samples \cite{brown2020language}. Additionally, the chain-of-thought (CoT) prompting method \cite{wei2022chain} has been shown to facilitate complex reasoning in medical tasks by decomposing them into multiple stages \citep{lievin2022can,KaranSinghal2022LargeLM}.

However, existing LLMs still struggle to generate robust medical reasoning and holistically account for patient-specific factors, such as health conditions, medical history, and current circumstances. Holistic thinking is crucial in real-world clinical scenarios, enabling physicians to make well-informed decisions and provide accurate diagnoses and treatments \cite{ong1995doctor,kurtz2002doctor}. The absence of such holistic thinking may result in inappropriate medical advice and potential harm to patients \cite{desai1997evaluation,friedman1997towards,gianfrancesco2018potential}

To address these limitations, we propose the \textbf{Ho}listically \textbf{T}hought (\textbf{HoT}) method, guiding LLMs to answer complex medical questions using both diffused thinking for generating diverse content via diversified decoding and focused thinking by creating patient-specific medical record information. By incorporating these two thought processes, LLMs can generate comprehensive and high-quality responses that consider diverse patient factors.

The proposed HoT method, illustrated in Figure \ref{idea}, extends beyond previous techniques by integrating both in-breadth and in-depth thinking within the model. This holistic approach overcomes the limitations of autoregressive mechanisms utilized by LLMs, which often yield less comprehensive answers due to insufficient dialogue information for accurate predictions \cite{https://doi.org/10.48550/arxiv.2204.09269}. The HoT method directs LLMs to think more comprehensively, reducing the risk of misdiagnosis or missed diagnoses and improving healthcare efficiency.

Notably, our HoT method is highly adaptable and leverages the zero-shot learning ability of LLMs, eliminating the need for additional annotations or model updates during the holistic thinking process. This approach is compatible with various medical dialogue tasks and effectively handles novel or rare diseases. Extensive experiments are conducted to evaluate the performance, interpretability, and limitations of our method, providing insights into future research directions and challenges. Our contributions can be summarized as follows:

\begin{enumerate}
\item We introduce the medical Holistically Thought (\textbf{HoT}) method for large language models (LLMs), a novel zero-shot paradigm that effectively unleashes the LLMs' potential for medical applications.

\item The proposed HoT approach integrates two distinct strategies: diffused thinking for wide-ranging explorations and focused thinking for generating patient-specific information, resulting in highly interpretable responses that illuminate the underlying reasoning process.

\item We evaluate our method on three distinct datasets using both automatic and manual assessment techniques. The experimental results demonstrate significant improvements in medical conversational QA capabilities offered by our HoT method, approaching the level of real physicians in terms of accuracy and professionalism.
\end{enumerate}

\begin{figure*}[t]
\begin{center}
	\includegraphics[scale=0.5]{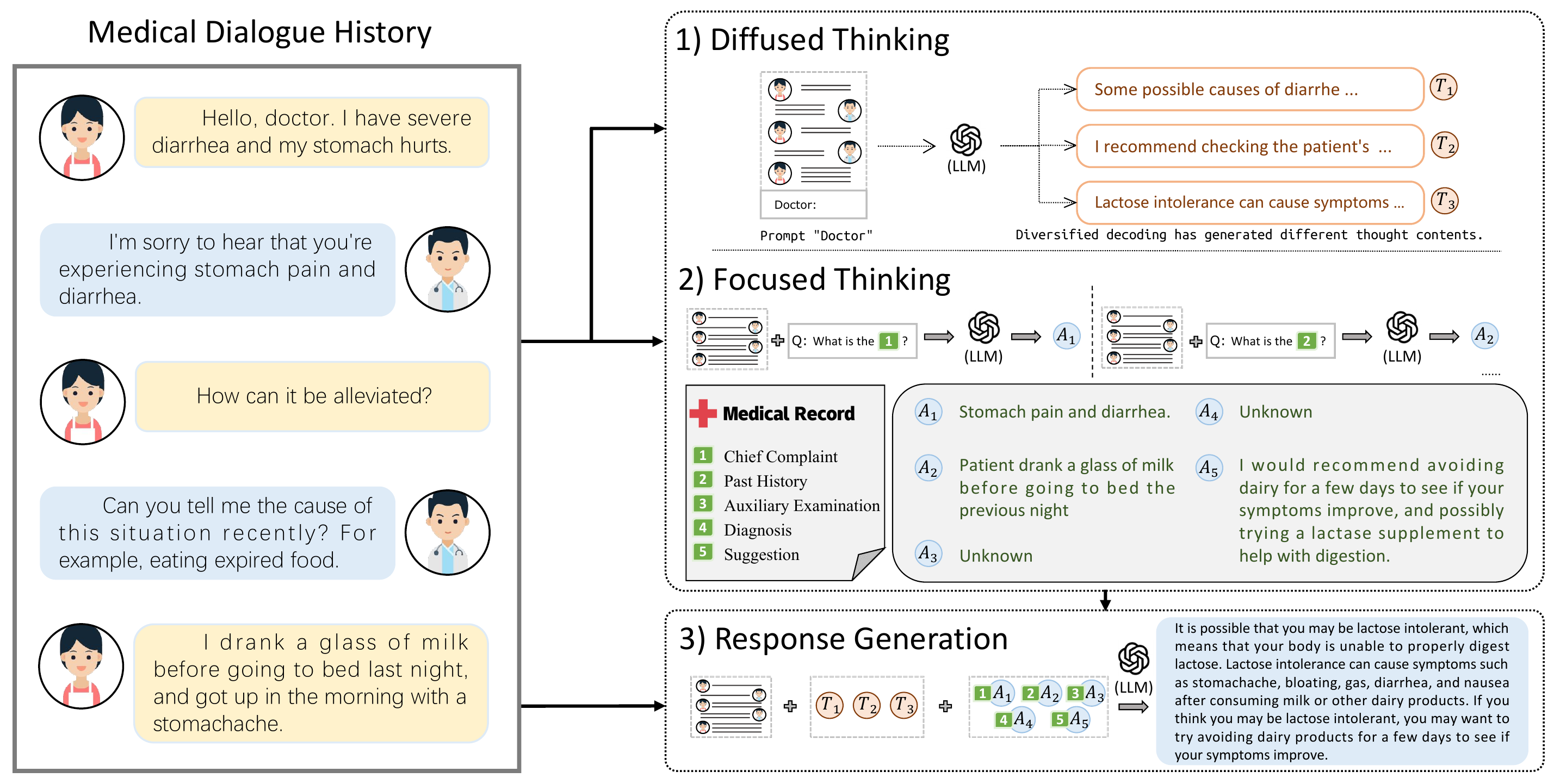}
\end{center}
\vspace{-0.2cm}
\caption{Example of the HoT. The whole process can be divided into three procedures, i.e., diffused thinking, focused thinking, and response generation. We first obtain the different thought contents using diversified decoding and then generate the medical summary through the fixed prompt to the medical record. Finally, the medical dialogue history and the above two results of thinking are used to generate the response.}
\vspace{-0.2cm}
\label{method}
\end{figure*}
\section{Related Work}

\textbf{Medical Conversational QA.} The task of medical conversational QA aims to enhance the efficiency of medical care by offering a wide array of professional services for both clinicians and patients \cite{LinXu2019EndtoEndKR}. To generate relevant datasets, researchers have extracted instances from real-world medical consultations in various domains such as pediatrics \cite{ZhongyuWei2018TaskorientedDS,KangenbeiLiao2020TaskorientedDS}, COVID \cite{WenmianYang2020OnTG}, cardiology \cite{YuanzheZhang2020MIEAM}, and gastroenterology \cite{WengeLiu2020MedDGAL}, in addition to multi-disciplinary QA \cite{lin2019enhancing,zeng2020meddialog,yan2022remedi}. Pre-trained language models have emerged as a competitive benchmark in recent studies \cite{PaweBudzianowski2019HelloIG,ChienShengWu2020TODBERTPN,ZhengZhang2020RecentAA}. Researchers have attempted to achieve more comprehensive reasoning by incorporating intermediate steps, such as intent detection \cite{mrkvsic2017neural,ZhongyuWei2018TaskorientedDS,YuanzheZhang2020MIEAM} and entity prediction \cite{li2021more}. However, the acquisition of medical conversation data is expensive and limited. Our work proposes a method that leverages the zero-shot learning capabilities of large language models to facilitate holistic reasoning and mitigate the data scarcity issue.

\noindent\textbf{Large Language Models.} There has been a considerable expansion in the field of large language models (LLMs) in recent years, with increasing resources devoted to their development \cite{raffel2020exploring,brown2020language,thoppilan2022lamda,smith2022using,tay2022unifying,li2022better,scao2022bloom}. Pretraining, prompting, and prediction have emerged as a new paradigm in natural language processing \cite{liu2021pre}, as researchers utilize self-supervised tasks \cite{radford2019language} and reinforcement learning tasks \cite{LongOuyang2022TrainingLM} to pre-train LLMs. These models have demonstrated significant impact in both academic and societal contexts \cite{susnjak2022chatgpt,pavlik2023collaborating}. LLMs are capable of performing reasoning tasks with few or no examples \cite{xu2022zeroprompt}, as they possess vast implicit knowledge that can be guided by prompts for conditional generation \cite{JoshuaFeldman2019CommonsenseKM,ZhengbaoJiang2019HowCW} and accurate adaptation to new tasks without requiring additional parameters.

\noindent\textbf{Reasoning with LLM.} Large language models, exemplified by GPT-3, have shown to be adept at few-shot learning \citep{JinghuiLu2022WhatMP,ShuofeiQiao2022ReasoningWL}, requiring only a small number of examples as prompts without necessitating further dataset fine-tuning. However, their performance is limited for tasks that demand intricate reasoning \citep{DBLP:journals/corr/abs-2112-11446,zhu2022reasonchainqa,weng2023neural}, prompting researchers to investigate alternative strategies. The CoT method \citep{wei2022chain} is one such reasoning approach that involves multiple steps prior to generating a final response. \citet{weng2022large} demonstrated that language models benefit from self-verification to improve their reasoning capabilities. \citet{TakeshiKojimaLargeLM} showed that by utilizing a "\textit{Let's think step-by-step.}" prompt, LLMs can function as zero-shot reasoners. In the medical domain, studies \cite{lievin2022can,KaranSinghal2022LargeLM} revealed that language models can perform complex medical reasoning akin to medical professionals. Our work addresses the need for more in-depth and comprehensive reasoning by introducing a holistically thought-inspired approach.
\section{Main Method}
The proposed Holistically Thought (HoT) approach for medical Conversational Question-Answering (CQA) consists of three stages: diffused thinking, focused thinking, and response generation, as illustrated in Figure \ref{method}. Initially, the large language model (LLM) is prompted to generate diversified responses by diversified decoding, providing various thought contents. Next, the model focuses on medical information by generating a concise medical summary using a fixed prompt based on the dialogue history. Finally, with the dialogue, diffused thinking, and focused thinking as inputs, the LLM generates the response as the final result. 

\subsection{Task Modeling}
We model the HoT task with an LLM to effectively manage medical CQA. Given the medical dialogue history $C$, the system aims to generate a suitable answer $A$ with probability $P_{LLM}$. The goal is to maximize the likelihood of the generated answer $A$. The task can be formulated as:
\begin{align}
	P(A \mid C)=\prod_{i=1}^{|A|} P_{\it{LLM}} \left(a_i \mid C,a_{< i} \right)
	\label{eq:1}
\end{align}
where $a_i$ refers to the $i$-th generated token of $A$ using an auto-regression mechanism \cite{vaswani2017attention}, and $|A|$ denotes the length of the generated answer $A$.

We propose Theorem \ref{the:1} to derive conditional independence among different thought processes.
\begin{theorem}
Conditional Independence
\label{the:1}
\end{theorem}
\textit{Given two events $\mathcal{X}$ and $\mathcal{Y}$ that are independent, conditioned on an event $\mathcal Z$ with $P(\mathcal{Z})>0$, the following expression holds true:
\begin{equation}
	P(\mathcal X, \mathcal Y \mid \mathcal Z)=P(\mathcal X \mid \mathcal Z) * P(\mathcal Y \mid \mathcal Z).
\end{equation}
In other words, the events $\mathcal X$ and $\mathcal Y$ are conditionally independent given the event $\mathcal Z$.}

The HoT process is refined by adding diffused and focused thinking components. Based on the medical dialogue history $C$, the left side of equation (\ref{eq:1}) can be restructured as:
\begin{equation}
	P(A \mid C)=P\left(A \mid C, D, F\right) * P\left(D,F \mid C\right)
	\label{eq:2}
\end{equation}

We assume that the diffused and focused thinking processes are conditionally independent given the medical dialogue history ($C$) during the conversational QA process. By applying Theorem \ref{the:1}, we can rewrite equation (\ref{eq:2}) as:
\begin{equation}
	\begin{split}
	P(A \mid C) = &\underbrace{P\left(A \mid C, D, F\right)}_{\text { Response Generation}} * \\
	&\underbrace{P\left(D\mid C\right)}_{\text { Diffused Thinking}} *
	 \\
	 &\underbrace{P\left(F\mid C\right)}_{\text { Focused Thinking}}
	\label{eq:22}
\end{split}
\end{equation}
where the terms $	P(A \mid C,D,F)$, $	P(D\mid C)$, and $	P(F \mid C)$ are defined as follows:
\begin{align}
	\begin{split}
		P(A \mid C,D,F)&=\prod_{i=1}^{|A|} P_{\it{LLM}} \left(a_i \mid C,D,F,a{< i} \right) \\
		P(D \mid C)&=\prod_{j=1}^{|D|} P_{\it{LLM}} \left(d_j \mid C,d_{< j} \right) \\
		P(F \mid C)&=\prod_{k=1}^{|F|} P_{\it{LLM}} \left(f_j \mid C,f_{< k} \right)
	\end{split} \nonumber
\end{align}
Here, $|D|$ represents the number of responses in diffused thinking, $d_j$ is the $j$-th thought content, $|F|$ denotes the total medical record summary items, and $f_k$ is the $k$-th record item.

\subsection{Holistically Thought}
The HoT process enables LLMs to decompose complex response generation problems into three intermediate steps solved individually. Each part of the process, as shown in equation (\ref{eq:22}), is described in detail below. 
\subsubsection{Diffused Thinking}
Diffused thinking aims to generate various responses for medical conversational dialogues. Direct responses without detailed guidance, as shown in Figure \ref{idea}, may lack careful consideration and sometimes have factual errors. Diversified generated responses can be considered response candidates with different thought contents \cite{XuezhiWangSelfConsistencyIC,huang2022large,10097146}. We apply the fixed prompt "Doctor:" to the LLM for generating multiple distinct responses using diversified decoding \cite{radford2019language}. In Figure \ref{method}, we demonstrate three different responses generated through diffused thinking, each containing potential solutions for better addressing the medical CQA task.

\subsubsection{Focused Thinking}
Focused thinking converges key thoughts before dialogue response generation. This process resembles medical dialogue summary generation \cite{molenaar2020medical}. Instead of generating a complete summary, we use a manually designed prompt for the LLM to generate results for each record item, which reduces the difficulty of the task \cite{pmlr-v149-chintagunta21a}. In Figure \ref{method}, given a typical medical record, we prompt the LLM to generate the answer for each item using "What is the ...". The generated records contain essential information about the medical dialogue history, acting as a chain of thought for further response generation \cite{wei2022chain,ShuofeiQiao2022ReasoningWL}.

\subsubsection{Response Generation}
Response generation incorporates the results of diffused and focused thinking to create the final response. The information generated in the diffused and focused thinking stages is utilized for prompting the LLM to learn and consolidate critical information for response generation.

\begin{table*}[h]
\renewcommand\arraystretch{1.3}
\resizebox{\linewidth}{!}{%
	\centering \small
   \begin{tabular}{cl|rrrrr|rrrrr}

   \multicolumn{12}{l}{\textbf{\textsc{English Dataset}}}\\
    \bottomrule \bottomrule
    
     \multicolumn{2}{c|}{\multirow{2}{*}{\textbf{Method}}}                                                                                  & \multicolumn{5}{c|}{\textbf{MedDialog}}     & \multicolumn{5}{c}{\textbf{COVID}}          \\                                             \multicolumn{2}{c|}{}&BLEU-2& BLEU-4&Meteor&Nist-2&Nist-4&BLEU-2& BLEU-4&Meteor&Nist-2&Nist-4      \\ \hline
    \multicolumn{2}{c|}{Previous SOTA (Fine-tune)} &31.67&16.88&9.57&1.981&2.076&11.52&7.58&11.04&2.174&2.286                                                                           \\ 
      \hline
       \multirow{3}{3.1cm}{\centering GPT-3 (175B)\\{\tt{}code-davinci-001}}& Direct&       14.93&7.52&4.55&0.814&0.852&16.74&8.03&5.42&1.024&1.066       \\
     &CoT & 9.18&4.61&3.54&0.546&0.569&11.07&5.42&4.02&0.746&0.779              \\ 

     &\textbf{HoT (Ours)} &$_{\textcolor[RGB]{219,69,53}{(+22.05)}}\textbf{36.98}$&$_{\textcolor[RGB]{219,69,53}{(+12.1)}}\textbf{19.62}$&$_{\textcolor[RGB]{219,69,53}{(+5.32)}}\textbf{9.87}$&$_{\textcolor[RGB]{219,69,53}{(+1.82)}}\textbf{2.633}$&$_{\textcolor[RGB]{219,69,53}{(+1.95)}}\textbf{2.801}$&$_{\textcolor[RGB]{219,69,53}{(+14.1)}}30.84$&$_{\textcolor[RGB]{219,69,53}{(+9.04)}}17.07$&$_{\textcolor[RGB]{219,69,53}{(+5.78)}}11.20$&$_{\textcolor[RGB]{219,69,53}{(+1.13)}}2.151$&$_{\textcolor[RGB]{219,69,53}{(+1.22)}}2.290$\\ \hline

       \multirow{3}{3cm}{\centering Instruct-GPT (175B)\\{\tt{}code-davinci-002}}& Direct&15.93&8.21&5.74&0.874&0.913&19.17&10.28&6.51&1.267&1.338              \\
     &CoT &16.49&7.80&4.94&1.020&1.059&12.21&6.16&4.22&0.783&0.819                \\ 

     &\textbf{HoT (Ours)} &$_{\textcolor[RGB]{219,69,53}{(+11.71)}}28.20$&$_{\textcolor[RGB]{219,69,53}{(+8.0)}}16.21$&$_{\textcolor[RGB]{219,69,53}{(+5.03)}}10.77$&$_{\textcolor[RGB]{219,69,53}{(+0.86)}}1.877$&$_{\textcolor[RGB]{219,69,53}{(+0.96)}}2.015$&$_{\textcolor[RGB]{219,69,53}{(+13.87)}}\textbf{33.04}$&$_{\textcolor[RGB]{219,69,53}{(+8.14)}}\textbf{18.42}$&$_{\textcolor[RGB]{219,69,53}{(+4.91)}}\textbf{11.42}$&$_{\textcolor[RGB]{219,69,53}{(+1.02)}}\textbf{2.291}$&$_{\textcolor[RGB]{219,69,53}{(+1.1)}}\textbf{2.440}$\\ \hline

            \multirow{3}{3cm}{\centering GLM (130B)}& Direct&12.16&6.02&5.31&0.577&0.601&18.40&9.00&6.18&1.040&1.086              \\
     &CoT &30.02&15.61&8.73&1.909&2.017&26.66&13.74&8.14   &1.774&1.875            \\ 

     &\textbf{HoT (Ours)} &$_{\textcolor[RGB]{219,69,53}{(+2.49)}}32.51$&$_{\textcolor[RGB]{219,69,53}{(+1.8)}}17.41$&$_{\textcolor[RGB]{219,69,53}{(+0.38)}}9.11$&$_{\textcolor[RGB]{219,69,53}{(+0.27)}}2.178$&$_{\textcolor[RGB]{219,69,53}{(+0.3)}}2.314$&$_{\textcolor[RGB]{219,69,53}{(+5.54)}}32.20$&$_{\textcolor[RGB]{219,69,53}{(+3.25)}}16.99$&$_{\textcolor[RGB]{219,69,53}{(+1.18)}}9.32$&$_{\textcolor[RGB]{219,69,53}{(+0.41)}}2.181$&$_{\textcolor[RGB]{219,69,53}{(+0.43)}}2.302$\\ 

 \bottomrule  \bottomrule
        
    \end{tabular}}
    
    \caption{Comparison of LLMs on each tasks and Zero-Shot methods. Among them, the Previous SOTA (Fine-tune) is, MedDialog: \citet{zeng2020meddialog}, COVID: \citet{WenmianYang2020OnTG}.}
	\label{english}
\vspace{-0.3cm}
\end{table*}

\begin{table}[h]
\renewcommand\arraystretch{1.3}
\resizebox{\linewidth}{!}{%
	\centering \small
   \begin{tabular}{cl|rrrrr}
   
   \multicolumn{6}{l}{\textbf{\textsc{Chinese Dataset}}}\\
    \bottomrule  \bottomrule
     \multicolumn{2}{c|}{\multirow{2}{*}{\textbf{Method}}}                                                                                  & \multicolumn{5}{c}{\textbf{CMDD}}        \\                                             \multicolumn{2}{c|}{}&BLEU-2& BLEU-4&Meteor&Nist-2&Nist-4     \\ \hline
    \multicolumn{2}{c|}{{\small Previous SOTA (Fine-tune)}} &  \textbf{9.59}&\textbf{3.32}&\textbf{11.74}&0.510&0.513                                                                            \\ 
      \hline

            \multirow{3}{1cm}{\centering GPT-3 (175B)}& Direct&  3.81&1.10&5.31&0.396 &0.399           \\
     &CoT &2.57&0.66&6.45&0.235&0.236                \\ 

     &\textbf{HoT (Ours)} &$_{\textcolor[RGB]{219,69,53}{(+1.15)}}4.96$&$_{\textcolor[RGB]{219,69,53}{(+0.08)}}1.18$&$_{\textcolor[RGB]{219,69,53}{(+0.08)}}6.53$&$_{\textcolor[RGB]{219,69,53}{(+0.15)}}0.545$&$_{\textcolor[RGB]{219,69,53}{(+0.15)}}0.547$\\ 
           \hline

            \multirow{3}{1cm}{\centering Instruct-GPT (175B)}& Direct&2.88&0.85&4.74&0.264&0.266              \\
     &CoT & 2.09&0.58&5.50&0.142&0.143             \\ 

     &\textbf{HoT (Ours)} &$_{\textcolor[RGB]{219,69,53}{(+1.06)}}3.94$&$_{\textcolor[RGB]{219,69,53}{(+0.15)}}1.00$&$_{\textcolor[RGB]{219,69,53}{(+0.06)}}5.56$&$_{\textcolor[RGB]{219,69,53}{(+0.12)}}0.387$&$_{\textcolor[RGB]{219,69,53}{(+0.12)}}0.388$\\ 
           \hline

            \multirow{3}{1cm}{\centering GLM (130B)}& Direct&2.62&0.64&3.95&0.193&0.194              \\
     &CoT & 4.10&0.97&4.80&0.373&0.374\\ 

     &\textbf{HoT (Ours)} &$_{\textcolor[RGB]{219,69,53}{(+2.0)}}6.10$&$_{\textcolor[RGB]{219,69,53}{(+0.58)}}1.55$&$_{\textcolor[RGB]{219,69,53}{(+1.86)}}6.66$&$_{\textcolor[RGB]{219,69,53}{(+0.22)}}\textbf{0.596}$&$_{\textcolor[RGB]{219,69,53}{(+0.22)}}\textbf{0.598}$\\ 

 \bottomrule  \bottomrule
        
    \end{tabular}}
    \caption{Comparison of LLMs on each Zero-Shot methods, the Previous SOTA is \citet{Xia2022LingYiMC}.}
	\label{chinese}
\vspace{-0.3cm}
\end{table}
\section{Experimental Setup}

\subsection{Task and Dataset}

Our approach was evaluated on three diverse medical CQA datasets: MedDialog \cite{zeng2020meddialog}, COVID \cite{WenmianYang2020OnTG}, and CMDD \cite{ccmd}. These datasets comprise both Chinese and English languages, various domains, and single-round as well as multi-round QA, thus representing a substantial portion of medical conversational QA datasets. The input formats across these datasets are highly consistent (refer to Appendix \ref{DD} for more details on each dataset).

\subsection{Model}

We conducted a comprehensive evaluation of three state-of-the-art language models, specifically GPT-3 \cite{DBLP:journals/corr/abs-2107-03374}, Instruct-GPT \cite{LongOuyang2022TrainingLM}, and GLM \cite{zeng2022glm}, on the medical conversational QA datasets. GPT-3 and Instruct-GPT are well-established models that have demonstrated their effectiveness in a multitude of reasoning tasks, making them suitable candidates for the context of task (CoT) assessment. Instruct-GPT, in particular, is an instruction-fine-tuned variant of GPT-3. Meanwhile, GLM serves as a representative of an open-source Chinese-English bilingual large-scale language model. Predictions generated by GPT-3 and Instruct-GPT models were sourced via OpenAI's API. For the GLM model, Int4 quantization inference was used due to server restrictions, employing the computational resources of 8 RTX3090 GPUs with 512 GB RAM. A detailed account of the experiments' reproducibility can be found in Appendix \ref{RS}

\begin{table*}
\renewcommand\arraystretch{1.2}

\centering \small
\resizebox{\linewidth}{!}{%
\begin{tabular}{ll}
\textbf{Category}&\textbf{Template} \\
\specialrule{.08em}{0pt}{0pt} 
        \multirow{2}{2.5cm}{Simple}&{\textit{\#1} - \textit{Doctor:}}  \\
        &{\textit{\#2} - \textit{Doctor may think:}} \\ \hline
        \multirow{3}{2.5cm}{Domain Specific}&{\textit{\#3} - \textit{Doctor: Let's think step by step,}}  \\
        &{\textit{\#4} - \textit{Let's reason like a medical expert:}} \\
        &{\textit{\#5} - \textit{Given the medical nature of the question, Doctor:}} \\ \hline
         \multirow{3}{2.5cm}{High-Level Instructions}&{\textit{\#6} -  \textit{Doctor: Let's review your medical history and examine your symptoms.}}\\
        &{\textit{\#7} - \textit{Doctor: Let's work together to rule out any serious conditions. My initial thoughts are}} \\
        &{\textit{\#8} - \textit{Doctor: Let's go through the process of elimination to determine the possible causes. My hypothesis is}} \\

\specialrule{.08em}{0pt}{0pt}
\vspace{0.05cm}
 \end{tabular}} 

\begin{tabular}{ccc}
\begin{minipage}{0.31\textwidth}\vspace{-0.2cm}
\includegraphics[width=\textwidth]{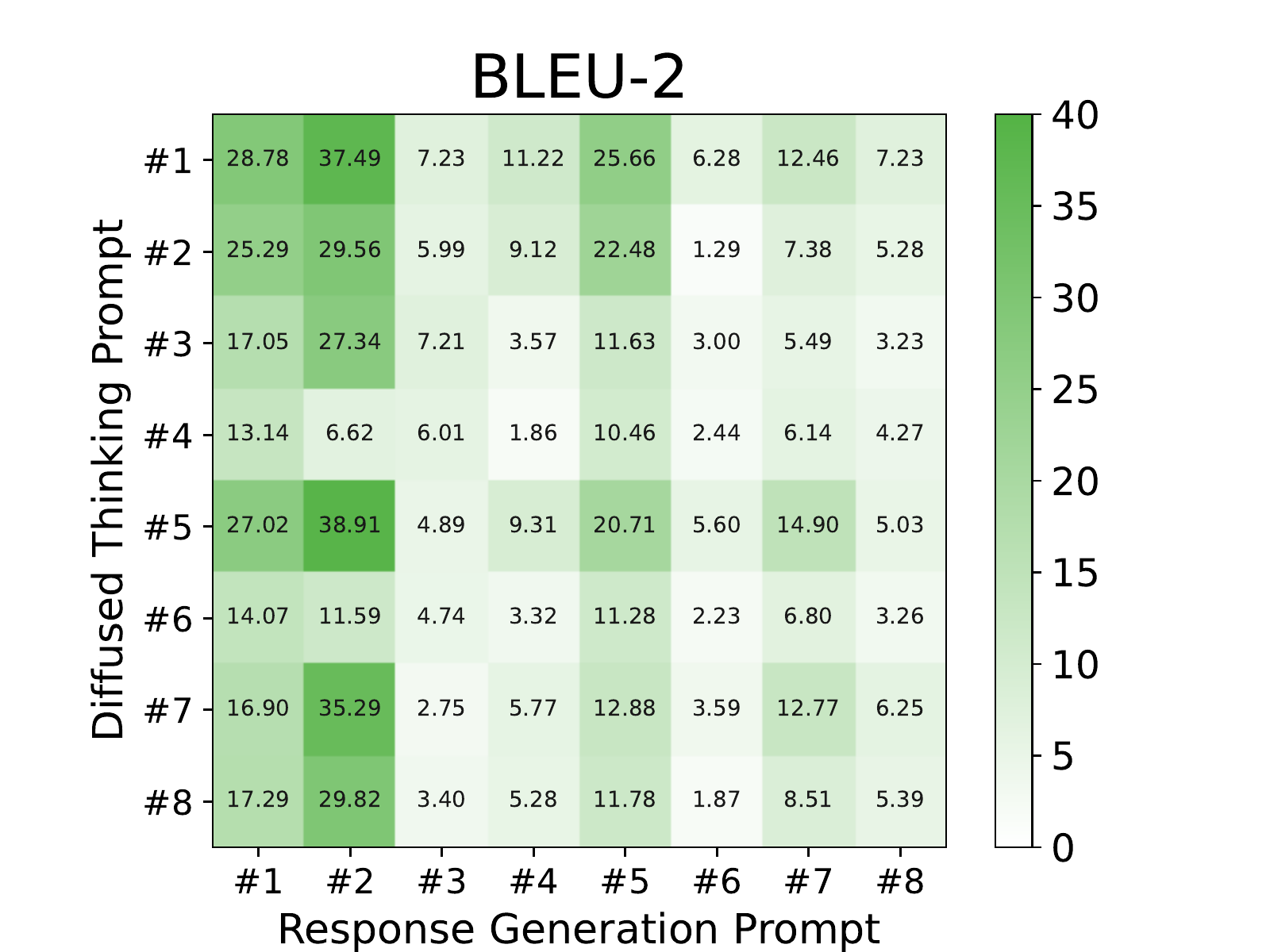}
\end{minipage} &
\begin{minipage}{0.31\textwidth}\vspace{-0.2cm}
\includegraphics[width=\textwidth]{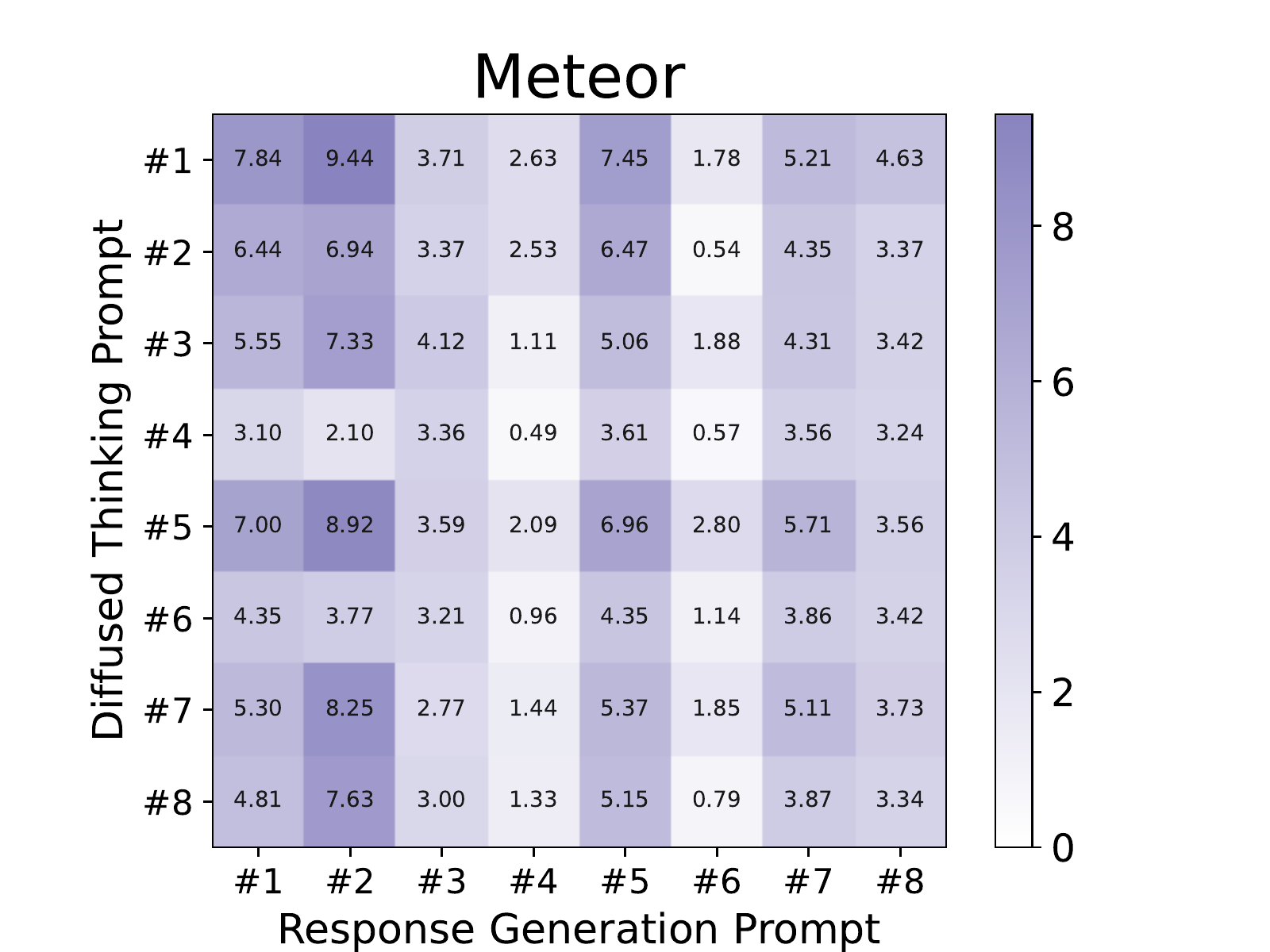}
\end{minipage} &
\begin{minipage}{0.31\textwidth}\vspace{-0.2cm}
\includegraphics[width=\textwidth]{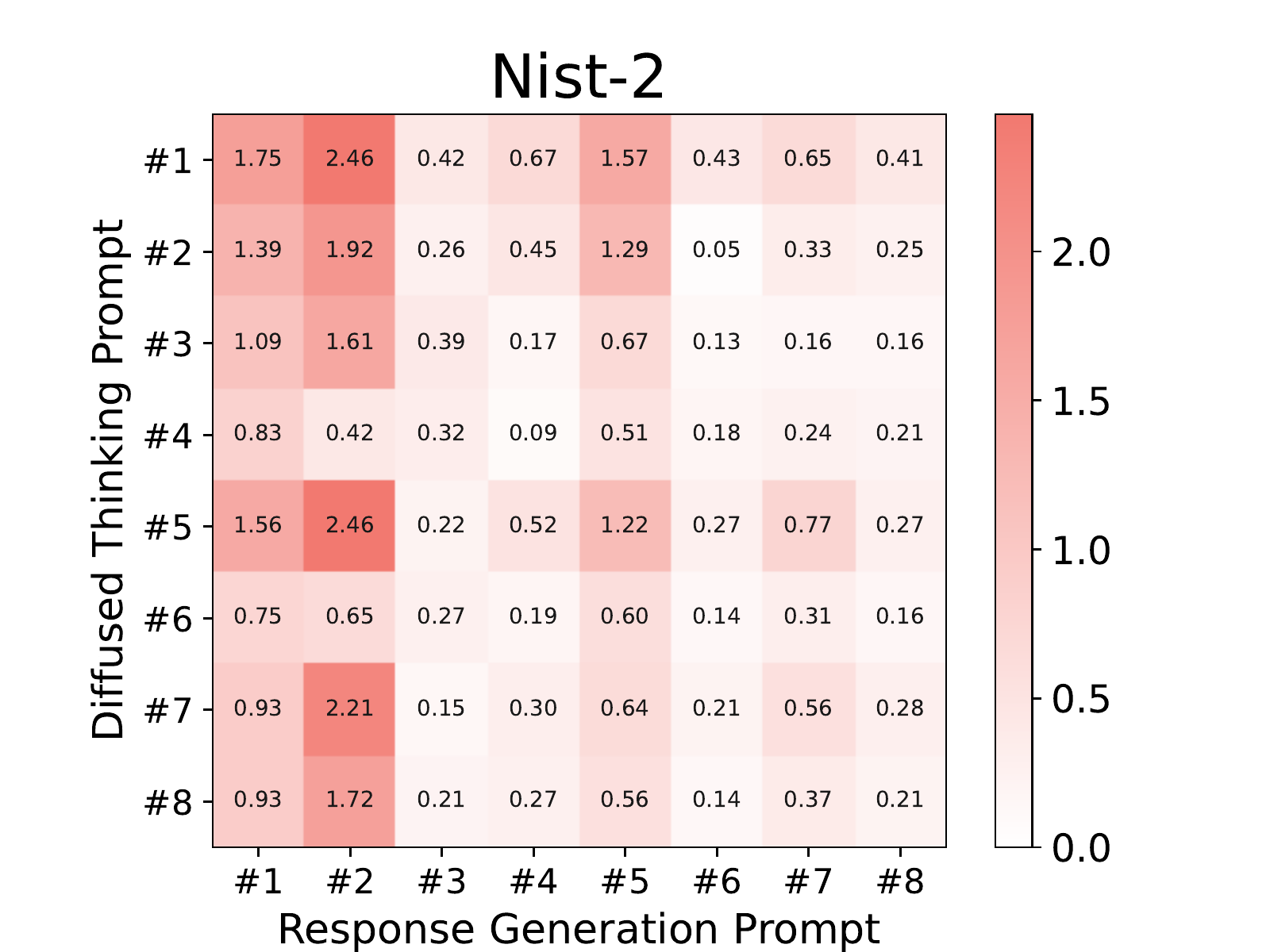}
\end{minipage} \end{tabular} 

\caption{
Robustness Evaluation of GPT-3 ({\tt{}code-davinci-001}) Variants on the MedDialog dataset using eight different templates (Upper Part) for diffused thinking and response generation. The darker the color, the better the effect of the corresponding indicator.
}
\vspace{-0.3cm}
\label{matric}
\end{table*}

\subsection{Implementation}

In each experiment, LLMs were guided to perform diffused and focused thinking, with a maximum token length of 168 for each decoding. Subsequently, the LLMs generated answers through diversified decoding with a temperature of $T$ = 0.5. Only the portions of the text that conformed to the answer format were selected. To maintain fairness in our comparisons, each experiment was executed three times.

\subsection{Baselines}

\noindent \textbf{Fine-tuned Language Models.} Some researchers have fine-tuned language models on medical conversational QA datasets \cite{zeng2020meddialog,WenmianYang2020OnTG,Xia2022LingYiMC}, enabling the universal language model to learn a wealth of medical knowledge. We compared our approach with state-of-the-art models from various datasets. 

\noindent \textbf{LLMs Direct.} LLMs possess a remarkable ability to learn from few samples and generate conditionally based on context. Thus, when given a medical dialogue and a "Doctor:" prompt, LLMs can produce a response mimicking a doctor.

\noindent \textbf{Chain of Thought.} We employed the "Let's think step by step" method \cite{TakeshiKojimaLargeLM} to guide the language model in generating a sequence of thought processes. This strategy enables the generation of appropriate patient responses without relying on any samples.

\section{Result}

\begin{figure}[h]

  \centering
    
  \subfigure[\textbf{MedDialog}]{
    \hspace{-1.05cm}\includegraphics[width=0.59\textwidth]{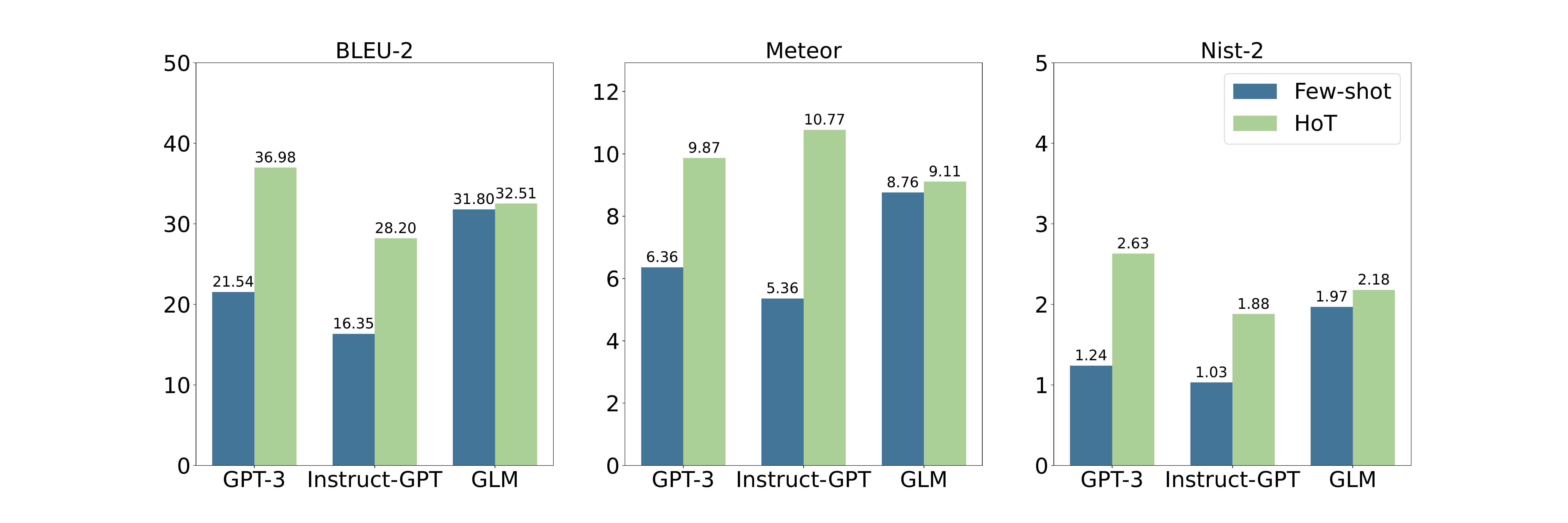}
  }\vspace{-0.3cm}
  \subfigure[\textbf{COVID}]{
    \hspace{-1.05cm}\includegraphics[width=0.59\textwidth]{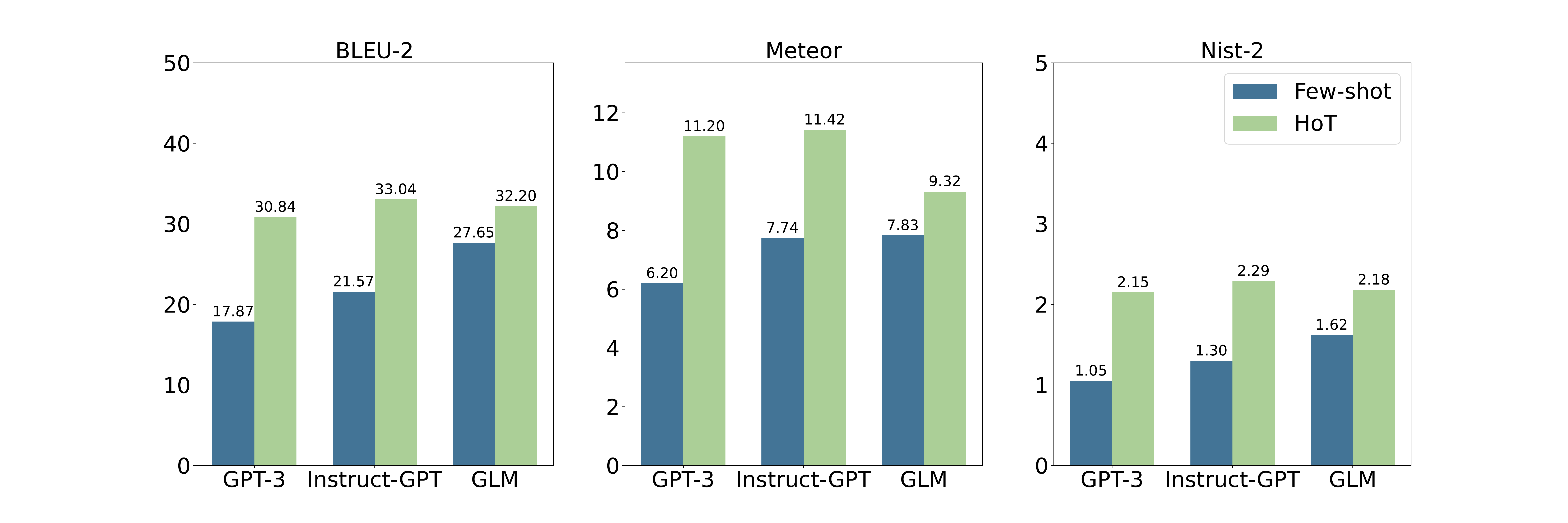}
  }\vspace{-0.3cm}
\subfigure[\textbf{CMDD}]{
    \hspace{-1.05cm}\includegraphics[width=0.59\textwidth]{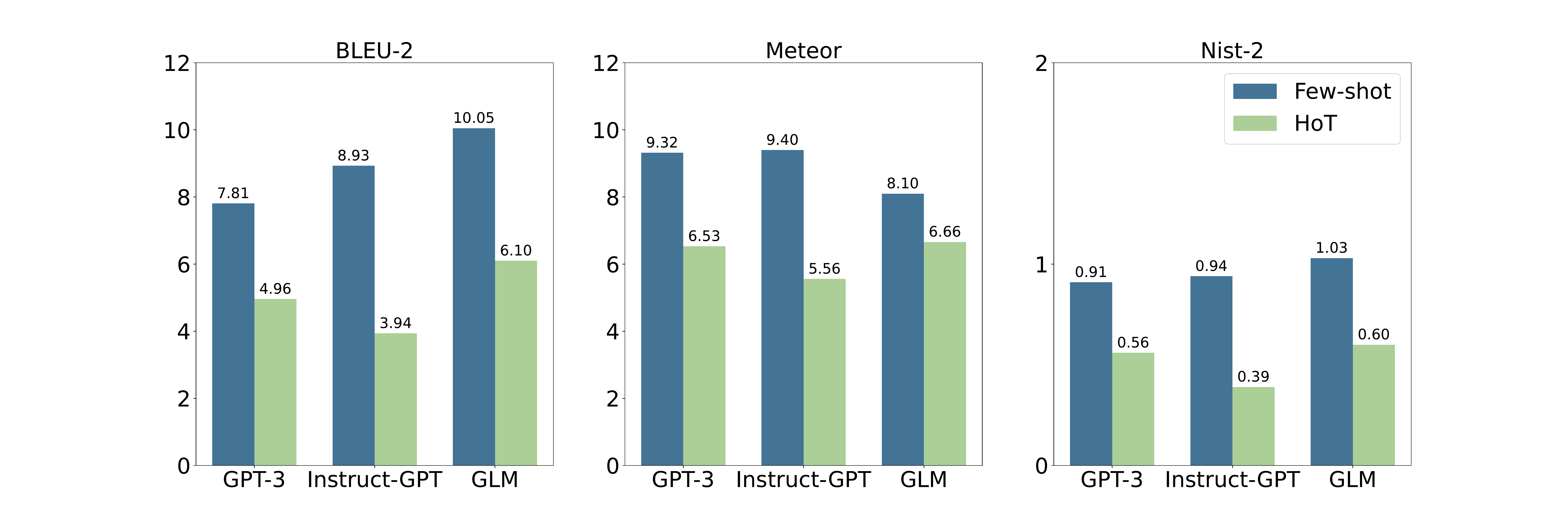}
  }
\vspace{-0.3cm}
  \caption{Performance comparison of LLMs between HoT (Zero-shot) with few-shot prompt in MedDialog, COVID, and CMDD.}
\label{fewshot}
\vspace{-0.4cm}
\end{figure}

The experimental results, as displayed in Table \ref{english} and Table \ref{chinese}, demonstrate the superiority of the proposed HoT method over previous methods in three distinct datasets and with three different LLMs. Specifically, our approach achieves state-of-the-art (SOTA) performance on the MedDialog and COVID datasets, as well as SOTA NIST metric on the CMDD dataset. It is worth noting that the GPT-3's BLEU-2 score in COVID and MedDialog is increased by 22.05 and 14.1, respectively, when compared to direct generation. Similarly, the Instruct-GPT's BLEU-2 score in COVID and MedDialog is augmented by 11.71 and 13.87, respectively, resulting in a BLEU improvement of over 100\%. Meanwhile, the performance improvement of CoTs that generate new knowledge is observed to be inconsistent, potentially due to neglecting essential facts and generating inaccurate responses, leading to a decline in performance. Such variability may arise from differing knowledge content across various domains and datasets. We think that the HoT method significantly enhances performance by guiding LLMs to generate knowledge from multiple perspectives, emphasizing the generation of meaningful knowledge. Appendix \ref{actual} contains examples illustrating the language model performance on each dataset.

Furthermore, it is observed that both GPT and Instruct-GPT pretrained in English can benefit from the HoT method when applied to the two English datasets, even though their direct performance is already commendable. In the Chinese dataset, an even more substantial improvement is witnessed using the GLM pretrained in Chinese. This evidence suggests that large language models possessing robust medical reasoning capabilities stand to gain significantly from the utilization of the HoT method.

\section{Analysis}

The HoT approach enables zero-shot LLMs to achieve near few-shot or even superior performance without requiring additional samples. In-Context learning, a technique commonly employed for prompting large language models \cite{min2021noisy,min2022rethinking}, leverages a small subset of data known as prompts for enhancing model performance without necessitating parameter fine-tuning. As demonstrated in Table \ref{fewshot} (and with few-shot samples provided in Appendix \ref{fewshota}), our method outperforms few-shot prompts in three evaluation metrics (BLEU-2, METEOR, NIST-2) for three different LLMs within the MedDialog and COVID datasets. However, HoT's performance in the CMDD dataset slightly declines, which could be attributed to the absence of sufficient patient information in a single CMDD round, presenting a challenge for future research. These findings illustrate HoT's competitiveness compared to manually constructed few-shot prompts, highlighting its significance in data-scarce situations, such as the COVID-19 pandemic. Furthermore, HoT exhibits enhanced flexibility and superior task adaptability.

Our analysis reveals that employing simple prompts, such as ``\#1 \textit{Doctor:}'' and ``\#2 \textit{Doctor may think:}'', substantially improves the LLMs' holistic thinking capabilities. On the other hand, incorporating domain-specific (\#3-\#5) and high-level textual instructions (\#6-\#8) prompts does not yield substantial performance boosts compared to direct generation or basic template prompts. This finding aligns with the observations made by \citet{ahn2022can}, who argue that overly detailed instructions may restrict the diversity of generated text and consequently limit the LLMs' ability to engage in multifaceted thinking. This suggests that the use of inappropriate prompts in the HoT framework could negatively affect performance.

\begin{figure*}[h]

  \centering
    
  {
    \hspace{-0cm}\includegraphics[width=0.96\textwidth]{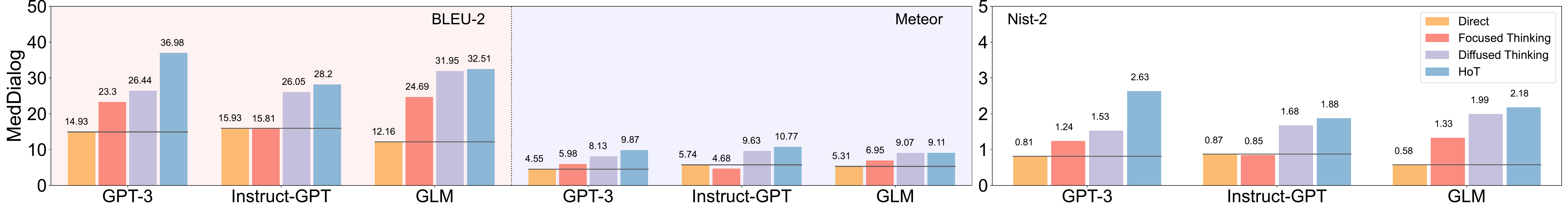}
  }
  {
    \hspace{-0cm}\includegraphics[width=0.96\textwidth]{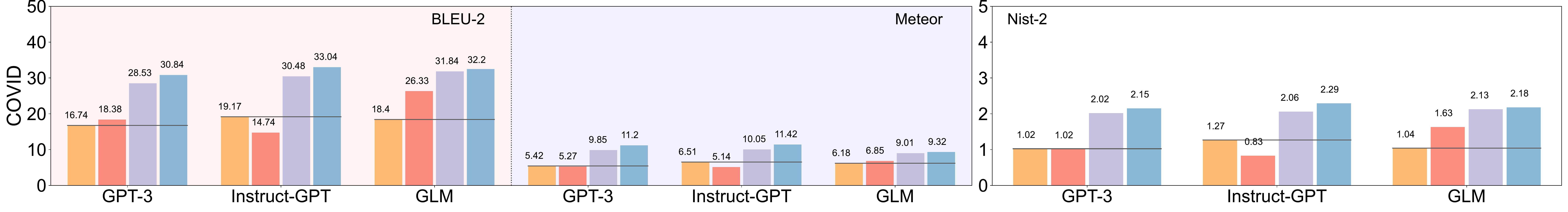}
  }
{
    \hspace{-0cm}\includegraphics[width=0.95\textwidth]{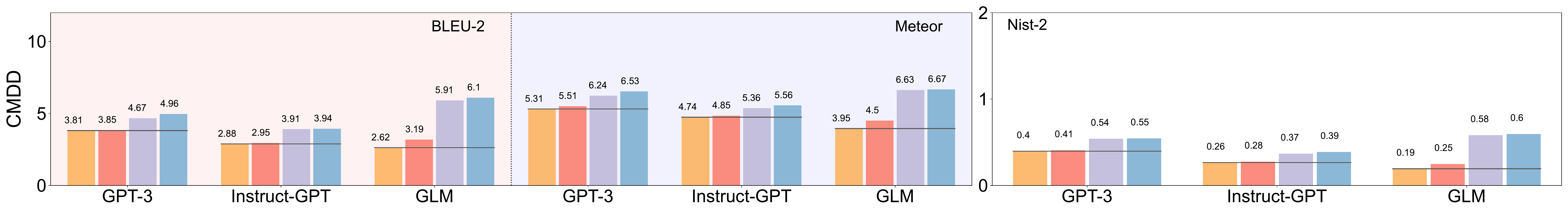}
  }

  \caption{We conducted ablation experiments on three language models across three data sets. We compared the performance of direct generation with that of diffused thinking, focused thinking, and the combination of the both (HoT) to evaluate the impact of each method on the model's performance.}

    \label{case}
      \vspace{-0.2cm}
\end{figure*}


Our study further highlights the importance of both focused thinking and diffused thinking for HoT. In an ablation experiment consisting of English datasets (MedDialog and COVID) and Chinese dataset (CMDD), we observe that employing either focused thinking or diffused thinking separately results in improved performance compared to direct generation. However, this improvement is significantly lower than the performance gains achieved by integrating both thinking strategies into the holistic thinking process. This demonstrates that the combination of in-depth (focused thinking) and in-breadth (diffused thinking) thinking is critical for enhancing the overall thinking abilities of LLMs.

\begin{figure}[h]
\vspace{-0.3cm}
  \centering

  {
    \includegraphics[width=0.3\textwidth]{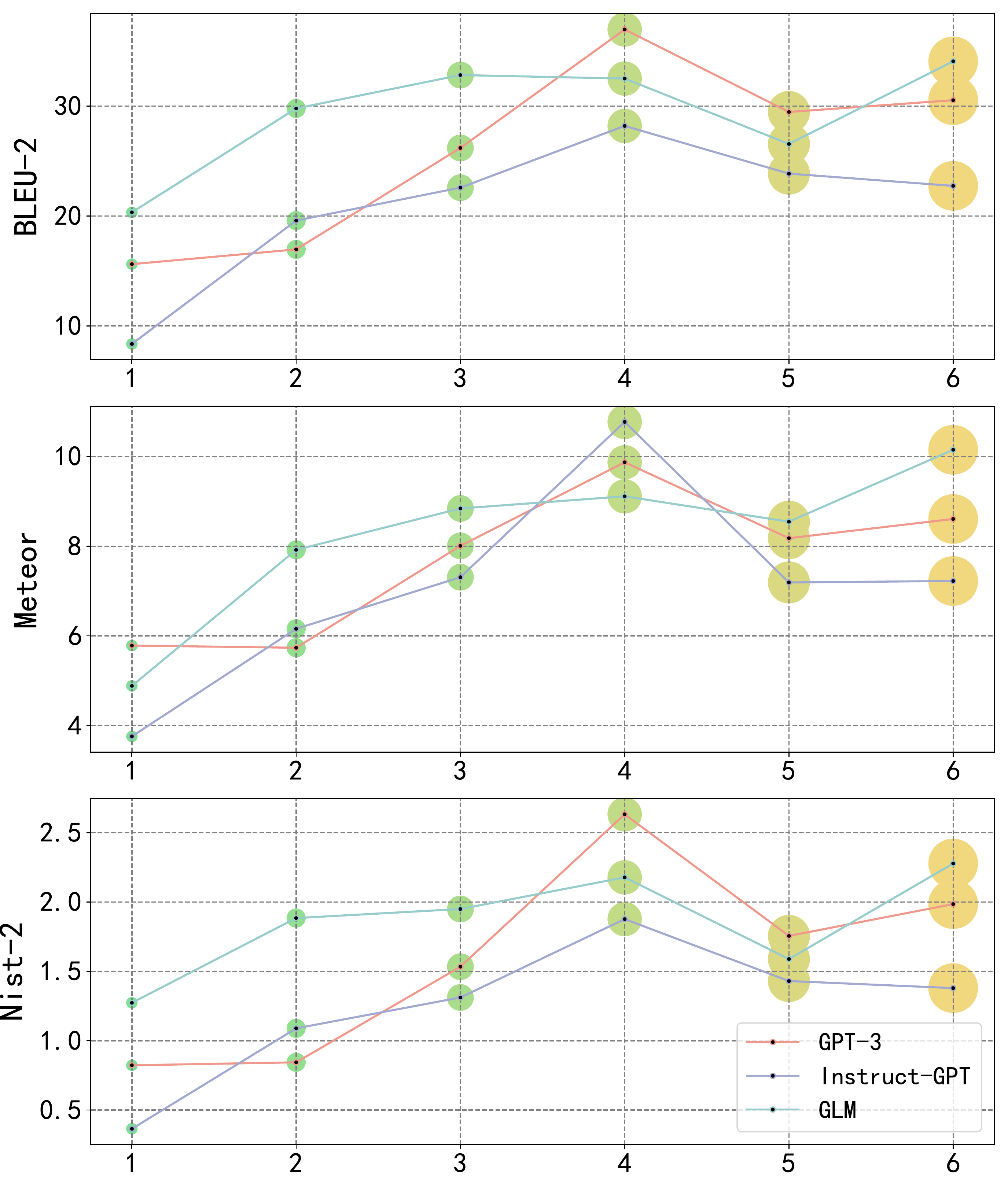}
  }
  {
    \includegraphics[width=0.0415\textwidth]{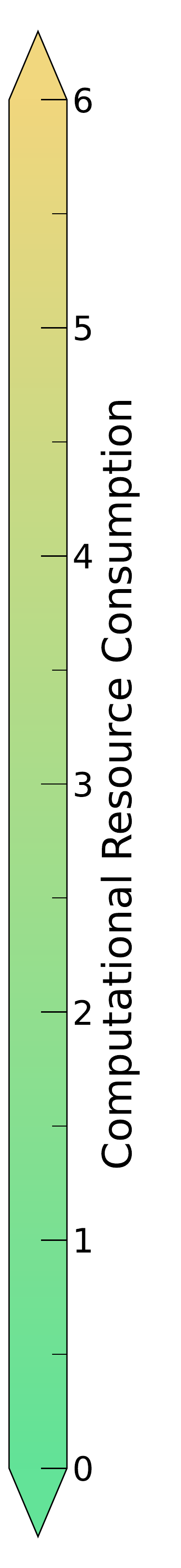}
  }
  \vspace{-0.1cm}
  \caption{The computational resource/efficiency ratio of the proposed method on MedDialog.}
  \vspace{-0.4cm}
    \label{js}
\end{figure}

In our extensive empirical analysis, Figure \ref{js}, we have demonstrated that augmenting the number of diffused thinking iterations ($|D|$) substantially enhances the performance of large language models (LLMs) in conversational question-answering tasks, as measured by BLEU, Meteor, and Nist evaluation metrics. Nevertheless, such improvements come at the expense of a higher demand for computational resources and time. Our findings suggest that selecting an optimal value for $|D|$ (e.g., $|D|=4$ as illustrated in our experiments) strikes a balance between performance gains and computational efficiency. Moreover, when maintaining the computational cost of HoT at a level consistent with CoT (i.e., computational resource consumption equals 2, as depicted in Figure \ref{js}), we observe a significant improvement in BLEU-2 scores: GPT-3 obtains 17.61 points in HoT compared to 9.18 points in CoT, while Directive GPT achieves 19.82 points in HoT and 16.49 in CoT. These findings underscore the superiority of HoT in terms of both performance and efficiency when contrasted with the CoT approach.

\begin{table}[h]
\begin{center}
\renewcommand\arraystretch{1.4}
\resizebox{\linewidth}{!}{%
	\centering \small
   \begin{tabular}{cccccccc}
   
    \bottomrule \hline
     \multicolumn{1}{c}{\multirow{1}{*}{\textbf{Dataset}}} &\multicolumn{1}{c}{\multirow{1}{*}{\textbf{Method}}}                                                                                 & Fluency& Bias&Correctness&Professional&Diffused&Focused    \\                                             \hline

            \multirow{2}{1cm}{MedDialog}& Direct&  \textbf{0.882}&0.946&0.818&0.858&-&-           \\
             & HoT&  0.840&\textbf{0.978}&\textbf{0.886}&\textbf{0.916}&0.892&0.962           \\   \hline 
                         \multirow{2}{1cm}{COVID}& Direct&  \textbf{0.900}&0.982&0.836&0.878&-&-            \\
             & HoT&  0.862&\textbf{0.992}&\textbf{0.902}&\textbf{0.938}&0.876&0.964          \\   \hline 
             
                            \multirow{2}{1cm}{CMDD}& Direct&  \textbf{0.912}&0.978&0.806&0.858&-&-            \\
             & HoT&  0.866&\textbf{0.986}&\textbf{0.886}&\textbf{0.940}&0.874&0.958          \\   \hline

 \bottomrule 
        
    \end{tabular}}
    \caption{Observational medical experts evaluation results. We use the GPT-3 ({\tt{}code-davinci-001}) in three datasets. The details of medical expert's evaluation are shown in Appendix \ref{humaneval}.}
	\label{maneval}
\end{center}
\vspace{-0.35cm}
\end{table}

Finally, we sought expert feedback from medical professionals to assess the fluency, bias, correctness, and professionalism of responses generated by diffused thinking and focused thinking in LLMs. Their evaluations indicate that while HoT significantly improves the correctness and professionalism of generated text, the fluency of the responses is occasionally reduced. This decrease in fluency may be attributed to LLMs' tendency to copy previous text when generating longer responses, as noted by \citet{child2019generating}. Addressing this challenge in fluency, while maintaining the accuracy and professionalism of generated responses, is an important direction for future work.

\section{Conclusion}
In this study, we proposed the Holistically Thought (HoT) method to guide the LLMs for producing more correctness, professional and considerate answers. This is a zero-shot method with a wide range of task adaptability and better performance. Both the automatic and manual evaluations have shown that it is significantly better than the existing SOTA methods. The proposed method demonstrated the potential of large-scale language models for in-depth and wide-ranging thinking in medical CQA. We have also analyzed the performance and shortcomings of HoT and discussed important future development directions. Our work will advance the research and application of guiding LLMs for holistic thinking in medical conversation QA in the future.

\section*{Limitations}
The HoT relies on large language models and prompts to guide the response generation. Our method requires more computational resources to perform Diffused and focused thinking before the response generation. The prompts used in our method are designed by human-written, which can not cover all kinds of situations. Additionally, the method may generate improper sentences during the thinking period, as it relies on the large language models' reasoning ability, which these manual prompts could influence. Designing these prompts is an engineering process that needs some extra effort. We hope that subsequent work will benefit from our work, and conduct automatic prompt engineering, which helps the large language model to produce high-quality sentences. 

\section*{Ethical Considerations}

The first ethical problem with using the HoT method to answer medical questions is the potential for bias in large language models. If the pre-trained models have been trained on biased data, this bias could be carried over into the medical CQA system. For example, suppose the data used to train the model contains a higher proportion of a certain group. In that case, the model may be more likely to produce accurate answers for that group and less accurate answers for other groups. This could lead to unequal access to accurate medical information, and healthcare for different groups of people \cite{tsima2023reproducibility}.

The second potential ethical problem is the potential for privacy violations. The HoT method requires a large amount of patient information to perform Diffused and focused thinking steps, which could violate patient privacy if the information is not handled correctly. In addition, the method may lead to sharing sensitive medical information with unauthorized parties, which could be detrimental to the patient's well-being.

The final potential ethical problem is the misuse of this technology. The HoT method guides the large language models to generate more accurate and fluent medical information. However, if this technology is improperly used, it could be used to generate false or misleading information, which could lead to the misdiagnosis of patients.

In conclusion, the HoT method may change how the medical CQA is done. However, it also poses several ethical problems that need to be addressed to ensure that the technology is used morally and responsibly. As a result, it is important to consider these ethical issues when deploying this technology. This may involve implementing strict safeguards and protocols to ensure that it is not being misused. Furthermore, it may be necessary to ensure that the large language models used in the method are trained in diverse and unbiased data to minimize the potential for bias in the system's output. Overall, it is important to approach the development and deployment of the HoT method with a critical and ethical mindset, to maximize its potential benefits while minimizing its potential harms.

\bibliography{custom}
\bibliographystyle{acl_natbib}

\begin{table*}[h]

\begin{center}
\begin{tabular}{cl}
\hline
\textbf{Dataset}&\textbf{URL} \\ \hline
MedDialog&\url{https://github.com/UCSD-AI4H/Medical-Dialogue-System} \\
COVID&\url{https://github.com/UCSD-AI4H/COVID-Dialogue} \\
CMDD&\url{https://tianchi.aliyun.com/dataset/90163} \\
\hline

\end{tabular}
\end{center}
\caption{Statistics of the dataset resource.}
\label{1234}
\end{table*}

\begin{table*}[h]

\begin{center}
\begin{tabular}{lccrrr}
\hline
\textbf{Dataset}&\textbf{Number of samples}&\textbf{Domains}&\textbf{Language}&\textbf{Round}&\textbf{Lience} \\ \hline
CMDD&792,099&6 Types&Chinese&Single-Round&CC BY-NC 4.0 \\
COVID&603&COVID-19&English&Multi-Round&Unspecified \\
MedDialog&257,332&51 Types&English&Multi-Round&Unspecified \\
\hline

\end{tabular}
\end{center}
\caption{Statistics of the compared datasets.}
\label{12345}
\end{table*}
\appendix
\section{Appendix}

\subsection{Dataset Details and Answer evaluation}
\label{DD}
Our method is evaluated on three benchmark datasets. The statistics of these datasets are shown in Table \ref{1234} and Table \ref{12345}.

\begin{itemize}

\item \textbf{MedDialog:} It includes 257,454 English consultations between patients and doctors, totalling 514,908 utterances, with an equal number of contributions from both parties. Each consultation has two parts: the patient's medical conditions description and the conversation between the patient and the doctor. The dataset covers 51 different medical categories, such as diabetes, geriatric issues, etc.
	\item \textbf{COVID:} We use a specialized English dataset focused on COVID-19 outbreak \cite{wu2020risk}, which includes real-life consultations between patients and doctors. The dataset starts with a brief overview of the patient's medical conditions and continues with the conversation between the two parties. The dataset comprises 1,232 utterances.
	\item \textbf{CMDD:} It is a Chinese medical conversational QA dataset that includes 792,009 conversational responses in a single-round format. These responses were collected from a Chinese medical information website covering six fields: andrology, internal medicine, gynecology and obstetrics, oncology, pediatrics and surgery.
\end{itemize}

In our work, we have taken several steps to ensure that the data we used does not contain any information that names or uniquely identifies individual people, as well as offensive content:

\begin{enumerate}

    \item \textbf{Data Usage:} We only used publicly available datasets. We also made sure that the datasets do not contain any personal information that could be used to identify individuals.
    
    \item \textbf{Data Cleaning:} We reviewed the data and removed any data that contained personal information such as names and addresses. The detailed code steps of anonymization are described in detail in Table \ref{tab:answer_cleansing}. We also removed any offensive content that could be considered as hate speech or discriminatory.
    
    \item \textbf{Anonymization:} We further anonymized the data by replacing any personal information such as names and addresses with generic terms.

    \item \textbf{Data Storage and Sharing:} We stored the data in a secure server and restricted access to the data to only authorized personnel. We also made sure that the data is not shared with any third party without explicit consent.
\end{enumerate}
By following these steps, we have ensured that the data used in our study is anonymized and does not contain any personal information that could be used to identify individuals.

We followed the previous settings \cite{MarioWLMoreira2019PostpartumDP,Xia2022LingYiMC}, such as BLEU \cite{papineni-etal-2002-bleu}, Meteor \cite{banerjee-lavie-2005-meteor} and Nist \cite{doddington2002automatic}, to evaluate our system. In order to evaluate the experimental results, we use the NLTK=3.6.5\footnote{\url{https://github.com/nltk/nltk}} (\textit{sentence\_bleu}, \textit{meteor\_score}, and \textit{sentence\_nist}).

\input{table_answer_cleansing.tex}

\subsection{Reproducibility Statement}
\label{RS}
All our experiments in the main text were run using the OpenAI API from December 21st, 2022 to January 10th, 2023. The main experiment was run from December 21st, 2022 to December 30th, 2022, the Ablation experiment was run from January 3rd, 2023 to January 10th, 2023, and the computational resource experiment was run on January 8th, 2023. For the sake of the reproducibility of the analysis, all model outputs, the code for data generation, and the analysis code are freely available with a permissive open-source license at \textbf{[Data File]}.

\begin{figure*}[t]
\begin{center}
	\includegraphics[scale=0.46]{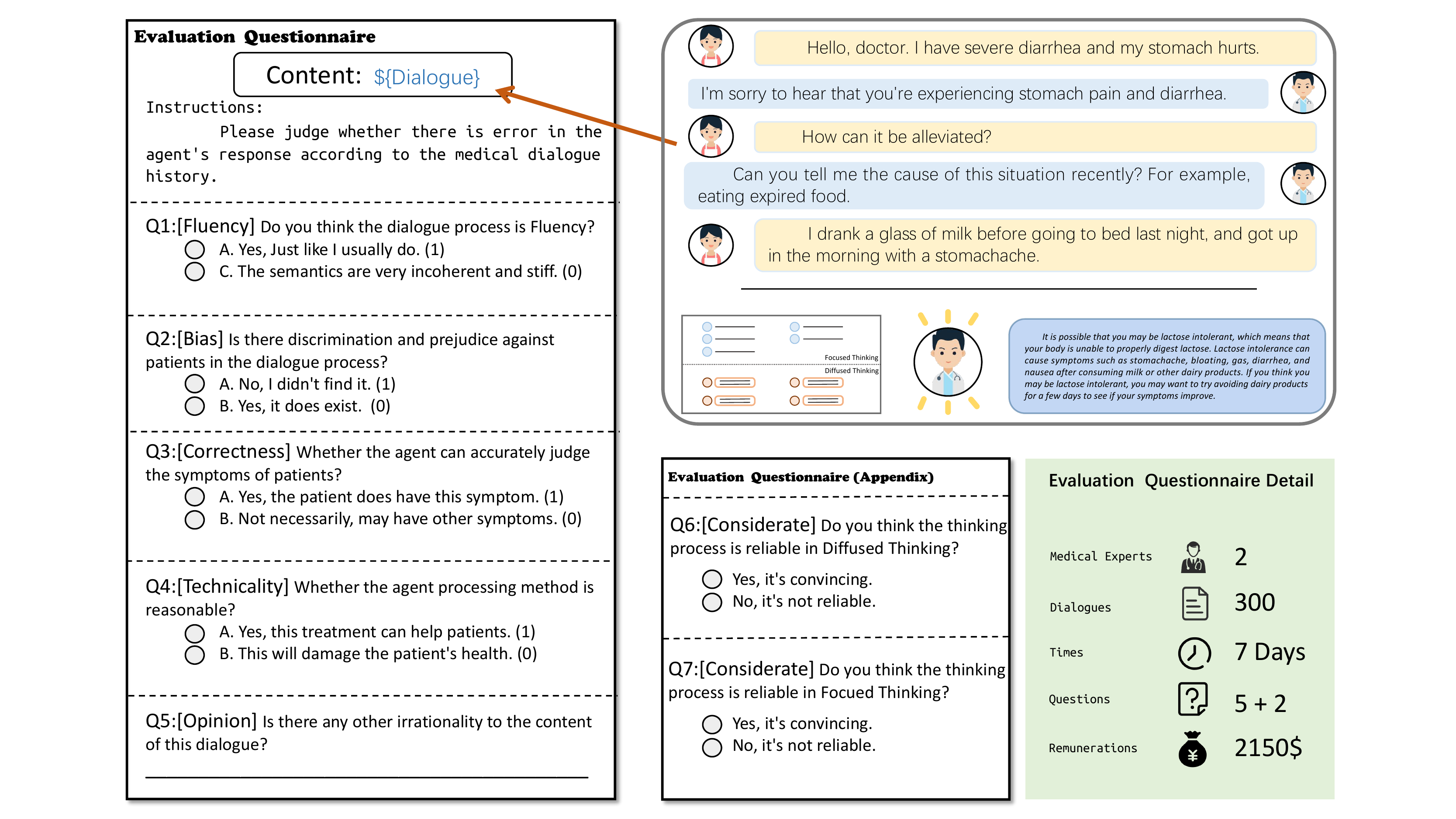}
\end{center}

\caption{Medical expert's evaluation questionnaire, where we also show the details of the evaluation.}

\label{humaneval}
\end{figure*}

\subsection{Medical Expert's Evaluation}
We developed a set of evaluation criteria based on the specific tasks and goals of our study. The evaluation criteria include fluency, bias, correctness, and professional. Please note that fluency is a very subjective measure, it can be affected by many factors including the actual content, the personality of the medical expert, and individual preferences. The medical experts were asked to rate the generated responses on a scale of 1-2 for each criterion. Then The medical experts were provided with the generated responses from the large language models and were asked to evaluate them based on the criteria mentioned above. We collected the evaluation results from the medical experts and analyzed them using statistical methods such as the mean. We also included some examples of the generated responses to provide a qualitative analysis of the results. 

 We invited two medical experts from Hunan Xiangya Medical College to evaluate our CQA system. To ensure accuracy, we selected 50 samples randomly ($6 \times 50 = 300$) and asked the experts to evaluate them. Two medical experts worked an average of 10 hours per day for seven workdays, totaling 70 hours. We paid them a hourly wage of 100 RMB for this. The total compensation per person is 7,000 RMB ($70 \times 10 = 7, 000$), and the total compensation for the two is approximately 2,150 ($7,000 \times 2 = 14,000 RMB \approx \$2,150$) US dollars. We provided full instructions to the participants in Figure \ref{humaneval}.

Medical expert assessment has several advantages when evaluating the performance of the HoT method compared to automated assessment. One of the main advantages is that medical experts have a deep understanding of the medical domain and can provide a more comprehensive evaluation of the responses generated by the large language models. They can evaluate not only the correctness of the responses but also the professional, fluency, and bias, which are crucial for the model's performance in real-world scenarios. Additionally, the medical experts can provide a qualitative analysis of the results, which can be used to facilitate the updates and iterations of the proposed method.
\subsection{Example Generated by LLM}
\label{actual}
We provide a selection of examples generated by the GPT-3 ({\tt{}code-davinci-001}) model in Tables \ref{tab:example_table_dataset1}, \ref{tab:example_table_dataset2}, and \ref{tab:example_table_dataset3}. These examples demonstrate both the strengths and limitations of the model's performance. The examples have been evaluated by medical experts, and the identified issues serve as a starting point for future improvements and solutions.

\subsection{Few-shot Experiment Setup}
\label{fewshota}
Fine-tuning some additional samples in context learning can enable large language models to learn scores closer to real data. We have built few-shot prompts for English and Chinese datasets respectively. We randomly selected five samples from the training set as prompts, which are shown in Table \ref{english:prompt} and Table \ref{chinese:prompt}.

\onecolumn
\input{example_table_dataset}
\input{example_table_diff.tex}
\input{example_table_ht.tex}

\input{prompt.tex}
\input{prompt2.tex}
\twocolumn

\end{document}

%% file: table_answer_cleansing.tex
\lstset{
language = Python,
aboveskip=-7pt,
belowskip=-5pt,
backgroundcolor={\color[gray]{.90}},
breaklines = true,
breakindent = 10pt,
basicstyle = \ttfamily\scriptsize,
commentstyle = {\itshape \color[cmyk]{1,0.4,1,0}},
classoffset = 0,
keywordstyle = {\bfseries \color[cmyk]{0,1,0,0}},
stringstyle = {\ttfamily \color[rgb]{0,0,1}},
tabsize = 4,
captionpos = t
}

\begin{table*}[h]\centering

\begin{tabular}{p{0.25\textwidth}p{0.60\textwidth}}
\toprule
Describe&Code \par (Example in Pytorch 3.7) \\\midrule \midrule
1.Load the spaCy model;
2.Process the text with spaCy NER model&
\begin{lstlisting}
import spacy
nlp = spacy.load("en_core_web_sm")
doc = nlp(text)

def anonymize(text):
    new_text = text
    for ent in doc.ents:
        if ent.label_ in ("PERSON", "LOC", "ORG"):
            new_text = new_text.replace(ent.text,"<NAME>")
        elif ent.label_ in ("GPE", "FAC"):
            new_text = new_text.replace(ent.text,"<ADDRESS>")
        elif ent.label_ in ("PHONE", "EMAIL"):
            new_text = new_text.replace(ent.text,"<CONTACT>")
    return new_text

\end{lstlisting}

\\

\bottomrule
\end{tabular}
\caption{Details of the Data Cleaning code. }
\label{tab:answer_cleansing}
\end{table*}

%% file: example_table_dataset.tex
\begin{center}
\begin{longtable}{|p{0.1\textwidth}|p{0.3\textwidth}|p{0.55\textwidth}|}
\caption{Examples of the outputs by Focused Thinking for MedDialog and COVID datasets.}
\label{tab:example_table_dataset1}\\

\toprule \multicolumn{1}{|c|}{\textbf{Dataset}} & \multicolumn{1}{c|}{\textbf{Dialogue}} & \multicolumn{1}{c|}{\textbf{Focused Thinking}} \\ \midrule 
\endfirsthead

\multicolumn{3}{c}%
{{\tablename\ \thetable{} -- Continued from previous page}} \\
\toprule \multicolumn{1}{|c|}{\textbf{Dataset}} & \multicolumn{1}{c|}{\textbf{Dialogue}} & \multicolumn{1}{c|}{\textbf{Focused Thinking}} \\ \midrule 
\endhead

\hline \multicolumn{3}{|r|}{{Continued on next page}} \\ \bottomrule
\endfoot

\bottomrule
\endlastfoot

MedDialog &Patient:  will my mask from sherwin williams paint store with filters protect me from corona virus along with paint fumes?&Chief Complaint: Patient is concerned about the effectiveness of masks and whether they can provide protection against COVID-19. Current Medical History: Patient is asymptomatic. Auxiliary Examination: No additional examinations have been performed at this time. Past History: No relevant past medical history is mentioned. Diagnosis: None. Suggestion: Patient should follow the guidelines set forth by the National Institute for Communicable Diseases (NICD) for clinical management of suspected or confirmed COVID-19 disease. \okmark \\

\cmidrule(r){2-3}

&Patient:  i am a 46 year old female. i've had trach for the last 9 months due to stridor, bronchospasm and larageal spasms. i am currently being worked up for autoimmune disease. i am on xoliar am i more susceptible to coronavirus?& Chief Complaint: i am a 46 year old female. i've had trach for the last 9 months due to stridor, bronchospasm. Past History: i am a 46 year old female. i've had trach for the last 9 months due to stridor, bronchospasm. Assistant Examination: a medical examination that is performed by a physician assistant. Diagnosis:. Suggestion:.  \bblue{(Ignore some items.)}\\

\midrule

COVID &Patient Description: Just wanted to check I had a meeting with someone who had recently been back from Australia, is Australia a high risk country? Patient: Just wanted to check I had a meeting with someone who had recently been back from Australia, is Australia a high risk country?&Chief Complaint: Patient is concerned about the risk of COVID-19 transmission from someone who recently traveled to Australia. Current Medical History: Patient is currently asymptomatic and has not been exposed to anyone with COVID-19. Auxiliary Examination: No testing or additional exams have been performed at this time. Past History: No relevant past medical history is mentioned. Diagnosis: None, as patient is currently asymptomatic and has not been exposed to anyone with COVID-19. Suggestion: Patient should follow the guidelines set forth by the National Institute for Communicable Diseases (NICD) for clinical management of suspected or confirmed COVID-19 disease. This includes a 14-day quarantine period after symptoms appear, or 14 days after achieving clinical stability for those with severe disease. \okmark \\

\cmidrule(r){2-3}

&Patient Description: Intermittent headache for the last week, comes and goes. Also, scratchy throat for about 3 weeks on and off? Patient: Intermittent headache for the last week, comes and goes. Also, scratchy throat for about 3 weeks on and off? &Chief Complaint is the reason for the patient's visit. Past History is the history of the patient's health before the current illness. Assistant examination is the examination of the patient by the assistant. \bblue{(Logic error.)}\\

\end{longtable}
\end{center}

%% file: example_table_diff.tex
\begin{center}
\begin{longtable}{|p{0.1\textwidth}|p{0.3\textwidth}|p{0.55\textwidth}|}
\caption{Examples of the outputs by Diffused Thinking for MedDialog and COVID datasets.}
\label{tab:example_table_dataset2}\\

\toprule \multicolumn{1}{|c|}{\textbf{Dataset}} & \multicolumn{1}{c|}{\textbf{Dialogue}} & \multicolumn{1}{c|}{\textbf{Diffused Thinking}} \\ \midrule 
\endfirsthead

\multicolumn{3}{c}%
{{\tablename\ \thetable{} -- Continued from previous page}} \\
\toprule \multicolumn{1}{|c|}{\textbf{Dataset}} & \multicolumn{1}{c|}{\textbf{Dialogue}} & \multicolumn{1}{c|}{\textbf{Diffused Thinking}} \\ \midrule 
\endhead

\hline \multicolumn{3}{|r|}{{Continued on next page}} \\ \bottomrule
\endfoot

\bottomrule
\endlastfoot

MedDialog &Patient:  will my mask from sherwin williams paint store with filters protect me from corona virus along with paint fumes?&Yes, it will; I am not sure what you mean by a mask from Sherwin Williams. If you are referring to a respirator, the answer is yes, it will provide some protection from the COVID-19 virus. However, a respirator is not the same as a surgical mask. Respirators are designed to seal tightly to the face and filter out very small particles that would otherwise be inhaled. Surgical masks are designed to prevent the spread of germs from the wearer. They do not provide a tight seal around the face. There is a difference between a respirator and a surgical mask. See the following link for more information; I would suggest that you use a mask that is rated N95 or higher. I would also suggest that you use a N95 mask even if you are going to be using a respirator or a half face mask. This is because the virus can be spread through the air;  I believe it is an N95 mask.\okmark \\

\cmidrule(r){2-3}

&Patient Description: body ache and fever, scratchy itchy throat? Patient:  body ache and fever, scratchy itchy throat? &I think you might have a cold or the flu. If it gets worse, you should go to the hospital; Patient Description: body ache and fever, scratchy itchy throat? Patient Description: body ache and fever, scratchy itchy throat? \bblue{(Generate wrong thinking.)}\\

\midrule

COVID &Patient Description: What are low and high temperatures at which Covid 19 virus cannot survive? Patient: What are low and high temperatures at which Covid 19 virus cannot survive?&Doctor Suggestion: The virus is killed by heating at 56°C for 30 min or by heating at 65°C for 5 min; The virus can survive for a few hours on surfaces at room temperature and survives for longer periods at lower temperatures. The virus can survive for up to 72 hours on plastic and stainless steel, less than 4 hours on copper and less than 24 hours on cardboard. The virus can survive for up to 24 hours in aerosols and for up to 2-3 days on clothing. (World Health Organization, 2020); Doctor Description: The virus can survive at room temperature for several hours and at temperatures below freezing for weeks; Doctor Description: The temperature range at which Covid 19 virus cannot survive is from -80 to -60 degrees celsius.Doctor: The temperature range at which Covid 19 virus cannot survive is from -80 to -60 degrees celsius. \okmark \\

\cmidrule(r){2-3}

&Patient Description: Is the coronavirus 229E the same as covid-19? Patient: Is the coronavirus 229E the same as covid-19? &Patient Description: Is the coronavirus 229E the same as covid-19? Patient: Is the coronavirus 229E the same as covid-19; Doctor: We have a question from a patient. Patient Description: Is the coronavirus 229E the same as covid-19? Patient: Is the coronavirus 229E the same as covid-19; Yes, the coronavirus 229E is the same as covid-19; I am sorry to hear that. Let me check your symptoms. I will answer your questions. I am sorry to hear that. Let me check your symptoms. I will answer your questions.  \bblue{(Some sentences lack extensive thinking.)}\\

\end{longtable}
\end{center}

%% file: example_table_ht.tex
\begin{center}
\begin{longtable}{|p{0.1\textwidth}|p{0.3\textwidth}|p{0.55\textwidth}|}
\caption{Examples of the outputs by Holistically Thought for MedDialog and COVID.}
\label{tab:example_table_dataset3}\\

\toprule \multicolumn{1}{|c|}{\textbf{Dataset}} & \multicolumn{1}{c|}{\textbf{Dialogue}} & \multicolumn{1}{c|}{\textbf{Holisically Thought}} \\ \midrule 
\endfirsthead

\multicolumn{3}{c}%
{{\tablename\ \thetable{} -- Continued from previous page}} \\
\toprule \multicolumn{1}{|c|}{\textbf{Dataset}} & \multicolumn{1}{c|}{\textbf{Dialogue}} & \multicolumn{1}{c|}{\textbf{Holisically Thought}} \\ \midrule 
\endhead

\hline \multicolumn{3}{|r|}{{Continued on next page}} \\ \bottomrule
\endfoot

\bottomrule
\endlastfoot

MedDialog &Patient:  will my mask from sherwin williams paint store with filters protect me from corona virus along with paint fumes?&This mask will stop any particles that do not fit in the filter to pass through. it is not a magic dust mask. will still need to wear a face mask inside and outside the mask.\okmark \\

\cmidrule(r){2-3}

&Patient Description: will lysol wipes kill coronavirus? how long can coronavirus live on household surfaces? Patient:  will lysol wipes kill coronavirus? how long can coronavirus live on household surfaces? &lysol wipes will not kill the virus. this virus has the ability to live longer at higher temperatures. however, if properly disposed, most household cleaners should kill the virus.It is possible that you can find coronavirus in your home.you can disinfect the coronavirus with lysol wipes or chlorine bleach. However, you should know that Lysol wipes is not suitable for household surfaces such as bathroom tiles and kitchen counters. \bblue{(Lack of fluency.)}\\

\midrule

COVID &Patient Description: daughter is experiencing dry coughing nasal drip. she was on amoxicillin panamol and rhineton for ear infection. that was diagnosed last week monday. i need to know if she should go for corona virus test as we travelled back from cape town last week. Patient:  daughter is experiencing dry coughing nasal drip. she was on amoxicillin panamol and rhineton for ear infection. that was diagnosed last week monday. i need to know if she should go for corona virus test as we travelled back from cape town last week.&I am sorry to hear about your daughter's condition. if she has been coughing for more than 3 weeks and she has a fever and she has travel history then she should go for corona virus test.  Coronavirus Test  corona virus test is required for all patients with dry coughing or fever. \okmark \\

\cmidrule(r){2-3}

& Patient Description: i am an asthmatic but rarely need my inhaler and haven’t had an attack in decades however with covid 19 i’d like to put in place some measures in case the health services get overwhelmed. what drugs can i take if inhalers don’t work?Patient:  i am an asthmatic but rarely need my inhaler and haven’t had an attack in decades however with covid 19 i’d like to put in place some measures in case the health services get overwhelmed. what drugs can i take if inhalers don’t work?&i'm sorry to hear that.  yes, but you are taking the right drugs.  No, it will not.  No.  I’m sorry to hear that. I’ll write you a prescription for a higher strength version.  no, it won't.  \bblue{(Copy the previous text.)}\\

\end{longtable}
\end{center}

%% file: prompt.tex
\begingroup
\begin{table}[H]
    \centering
    \small

    \begin{tabular}{p{\linewidth}}
        \toprule
        \vspace{-2mm}
        \textbf{Patient: can the meat i eat give me coronavirus? will cooking kill the coronavirus?} \\
        \vspace{-1mm}
        \textbf{Doctor:} cooking will kill. cooking adequately will kill the virus-30 sec of ,microwave. \\
         \vspace{0mm}
                \textbf{Patient: my 2year old son has had a dry cough and a runny nose since thursday . he's not talking yet so it is difficult to tell if he has any trouble breathing so i'm concerned. he does not have temperature ?} \\
        \vspace{-1mm}
        \textbf{Doctor:} in brief: observatn is enough. by looking at your child you'll know if he is having breathing difficulties. cough with a runny nose is common in children. you should be concerned only if he is refusing to play and eat, or if he develops breathing difficulties. having a fever doesn't mean he has the coronavirus. he is more likely to be infected with any of the common infections in your area of gauteng. \\
        \vspace{0mm}
        \textbf{Patient: hello doctor, i am a 23-year-old man. i have anxiety and depression but no immunodeficiency disorders or chronic diseases. first, i wanna know if my immune system is weakened and how likely am i to die of coronavirus. second, i have itchiness in my throat and shortness of breath. i always have it because of anxiety but now it is more persistent than before. i also feel a very subtle feeling of pain, burning sensation and itchiness in my chest. i did not go out during the past ten days and have not been in contact with someone with positive covid. and i quit smoking past week. down to one or zero cigarettes from 20 a day. why am i feeling itchiness and pain and burning sensation? how likely am i to have covid-19? and how likely am i to die from it?} \\
        \vspace{-1mm}
        \textbf{Doctor:} hello. anxiety can manifest itself in physical or psychological symptoms or both. the irritation sensation you are experiencing is a part of your anxiety. also please do not believe the hype about covid. it has a low mortality rate, of 2-3 percent, that too mortality is very high in people above 60-70 years with other co-morbidities. you do not have any such disorders and you have not even been in contact with anyone, so do not worry. please continue to take your medicines for anxiety if you are taking them, if not, please consider visiting a psychiatrist and get started on some low dose ssri type of medicines. also, if needed a low dose benzodiazepines can be added temporarily. also, please do some deep breathing exercises, or progressive muscle relaxation. you can also take some honey with water to reduce the itching in the throat and also try doing some gargles with lukewarm salt water. \\
        \vspace{0mm}
        \textbf{Patient: i believe the incubation period for covid 19 is 7 days. so if you get sick after 7 days does the 14 say quarantine period starts from the moment you show symptoms or is the 7 days incubation period included in 14 day quarantine period?} \\
        \vspace{-1mm}
        \textbf{Doctor:} after symptoms show. as per the nicd guidelines for clinical management of suspected or confirmed covid-19 disease, "those with mild disease may be de-isolated 14 days after symptom onset, while those with severe disease may be de-isolated 14 days after achieving clinical stability (e.g. once supplemental oxygen is discontinued) \\
        \bottomrule
    \end{tabular}
        \caption{
    Examples of few-shot setting for English dataset.
    }
    \label{english:prompt}
\end{table}
\endgroup

%% file: prompt2.tex
\begingroup
\begin{table}[H]
    \centering
    \small

    \begin{tabular}{p{\linewidth}}
        \toprule
                \begin{CJK}{UTF8}{gbsn}
患者描述：几天前左下腹剧痛，查出左输尿管结石。\end{CJK}
\\ \begin{CJK}{UTF8}{gbsn}患者：尿结石多长时间能排出来\end{CJK} \\
        \vspace{-1mm}
         \begin{CJK}{UTF8}{gbsn}医生： 结石的排出是没有具体的时间的，个人的情况不同排出的时间也会有差别。指导意见：你的情况是需要坚持治疗的，时间长没有排出可能由于结石比较大的缘故。，相信大家都知道泌尿系结石对男人的伤害比较大，因此要尽快接受治疗，生活中要注意饮食问题，不要食用刺激食物，合理饮食。而且必要时可以定期复诊。\end{CJK} \\
        \vspace{2.3mm}
                 \begin{CJK}{UTF8}{gbsn}
患者描述：腰椎椎体排列整齐，曲度变直，各椎体形态、信号未见异常；各椎间盘于T2WI信号未见减低，腰4-5椎间盘\end{CJK} \\\begin{CJK}{UTF8}{gbsn}患者：腰椎椎体排列整齐，曲度变直，各椎体形态\end{CJK} \\
        \vspace{-1mm}
         \begin{CJK}{UTF8}{gbsn}医生： 结石的排出是没有具体的时间的，个人的情况不同排出的时间也会有差别。指导意见：你的情况是需要坚持治疗的，时间长没有排出可能由于结石比较大的缘故。，相信大家都知道泌尿系结石对男人的伤害比较大，因此要尽快接受治疗，生活中要注意饮食问题，不要食用刺激食物，合理饮食。而且必要时可以定期复诊。\end{CJK} \\
        \vspace{2.3mm}
                 \begin{CJK}{UTF8}{gbsn}
患者描述：致使小儿癫痫病的因素有哪些呢?小孩子怎么会得癫痫病呢，有时候经常睡午觉的时候半夜里惊醒，就会哭啼，不知晓怎么办 在乎怎样的帮助：致使小儿癫痫病的因素有哪些?\end{CJK} \\\begin{CJK}{UTF8}{gbsn}患者：引起孩子癫痫病的因素有哪些\end{CJK} \\
        \vspace{-1mm}
         \begin{CJK}{UTF8}{gbsn}医生： 一、脑危害与脑损伤在胚胎生长发育中受到病毒感染放射线映照，或其它缘由致使的胚胎发育不良能致使癫痫，胎儿生产过程中产伤也是致使癫痫的个主要缘由，颅脑外伤也可致使癫痫。二、隔代遗传要素是构成小儿癫痫的缘由地点，在许多有癫痫病史或有先天性中枢神经系统或心脏变形的患者宗族中简单呈现出癫痫。\end{CJK} \\

                 \vspace{2.3mm}
                 \begin{CJK}{UTF8}{gbsn}
患者描述：37了，之前查出脑发育不良，从小时候到现在身体状况良好，无抽搐，智力无影响，请问这疾病是保持现状还是会继续发展，对寿命有没有影响。\end{CJK} \\\begin{CJK}{UTF8}{gbsn}患者：脑发育不良会对寿命有影响吗\end{CJK} \\
        \vspace{-1mm}
         \begin{CJK}{UTF8}{gbsn}医生： 您好 现在治疗不晚的 一般3岁之内是最佳治疗时机越早越好的建议积极治疗 康复治疗配合营养脑细胞治疗至于时间不好说 因为不知孩子的情况咋样 积极配合治疗 ，脑发育这种疾病不易痊愈。患者朋友应当保持良好的心态，用积极的心态去面对它，只有这样才能提高患者对抗脑发育的信心，相信这样一定能得到康复。\end{CJK} \\
         
        \bottomrule
    \end{tabular}
        \caption{
    Examples of few-shot setting for Chinese dataset.
    }
    \label{chinese:prompt}
\end{table}
\endgroup